\documentclass{bmvc2k}

\usepackage{multirow}
\usepackage{float} 
\usepackage[]{xcolor}
\usepackage{arydshln}
\usepackage{cleveref}
\usepackage{soul}
\usepackage[abs]{overpic} 

\usepackage{amssymb}

\newcommand{\myparagraph}[1]{\smallskip\noindent\textbf{#1}\quad}
\newcommand{\overimg}[3][]{%
    \begin{overpic}[#1]{#2}%
      \put (0, 2) {%
        \setlength{\fboxsep}{2pt}%
        \colorbox{cyan!0!white}{%
          \scriptsize\sffamily\vphantom{y}%
          #3%
        }%
      }%
    \end{overpic}%
}

\title{Adapting MIMO video restoration networks to low latency constraints}

\addauthor{Valéry Dewil}{firstname.lastname@ens-paris-saclay.fr}{1}
\addauthor{Zhe Zheng\vspace{-1em}}{}{1}
\addauthor{Arnaud Barral\vspace{-1em}}{}{1}
\addauthor{Lara Raad}{lara.raad@upf.edu}{2}
\addauthor{Nao Nicolas}{firstname.lastname@fr.thalesgroup.com}{3}
\addauthor{Ioannis Cassagne\vspace{-1em}}{}{3}
\addauthor{Jean-michel Morel}{jeamorel@cityu.edu.hk}{4}
\addauthor{Gabriele Facciolo\vspace{-1em}}{}{1} 
\addauthor{Bruno Galerne}{bruno.galerne@univ-orleans.fr}{5}
\addauthor{Pablo Arias}{pablo.arias@upf.edu}{6}

\addinstitution{Université Paris-Saclay,ENS Paris-Saclay, CNRS, Centre Borelli, France}
\addinstitution{Facultad de Ingeniería, Universidad de la República,  Uruguay}
\addinstitution{Thales LAS, France}
\addinstitution{City University of Hong Kong, Hong Kong}
\addinstitution{
Université d'Orléans, Institut Denis Poisson, 
France}
\addinstitution{Engineering Department, \\Pompeu Fabra University, Spain}

\runninghead{Dewil et al.}{Adapting MIMO networks to low latency constraints}

\def\etal{\emph{et al}\bmvaOneDot}
\definecolor{brickred}{rgb}{0.8, 0.25, 0.33}

\newcommand{\pa}[2]{{\color{gray}#1}{\color{cyan}#2}}

\newcommand{\gf}[2]{{\color{gray}#1}{\color{purple}#2}}
\newcommand{\vd}[2]{{\color{gray}#1}{\color{olive}#2}}
\newcommand{\lr}[2]{{\color{gray}#1}{\color{orange}#2}}
\newcommand{\zz}[2]{{\color{gray}#1}{\color{teal}#2}}

\newcommand{\nada}[1]{}
\definecolor{darkgreen}{rgb}{0, 0.5,0}

\begin{document}

\maketitle

\begin{abstract}
MIMO (multiple input, multiple output) approaches are a recent trend in neural network architectures for video restoration problems, where each network evaluation produces multiple output frames. 
The video is split into non-overlapping stacks of frames that are processed independently, resulting in a very appealing trade-off between output quality and computational cost. 
In this work we focus on the low-latency setting by limiting the number of available future frames. 
We find that MIMO architectures suffer from problems that have received little attention so far, namely 
(1) the performance drops significantly due to the reduced temporal receptive field, 
particularly for frames at the borders of the stack, 
(2) there are strong temporal discontinuities at stack transitions which induce a step-wise motion artifact.
We propose two simple solutions to alleviate these problems: recurrence across MIMO stacks to boost the output quality by implicitly increasing the temporal receptive field, and overlapping of the output stacks to smooth the temporal discontinuity at stack transitions.
These modifications can be applied to any MIMO architecture. 
We test them on three state-of-the-art video denoising networks with different computational cost.
The proposed contributions result in a new state-of-the-art for low-latency networks, both in terms of reconstruction error and temporal consistency.
As an additional contribution, we introduce a new benchmark consisting of drone footage that highlights temporal consistency issues that are not apparent in the standard benchmarks.
\end{abstract}

\begin{figure}[!ht]
    \centering
    \begin{tabular}{cc}
        \includegraphics[width=0.47\linewidth]{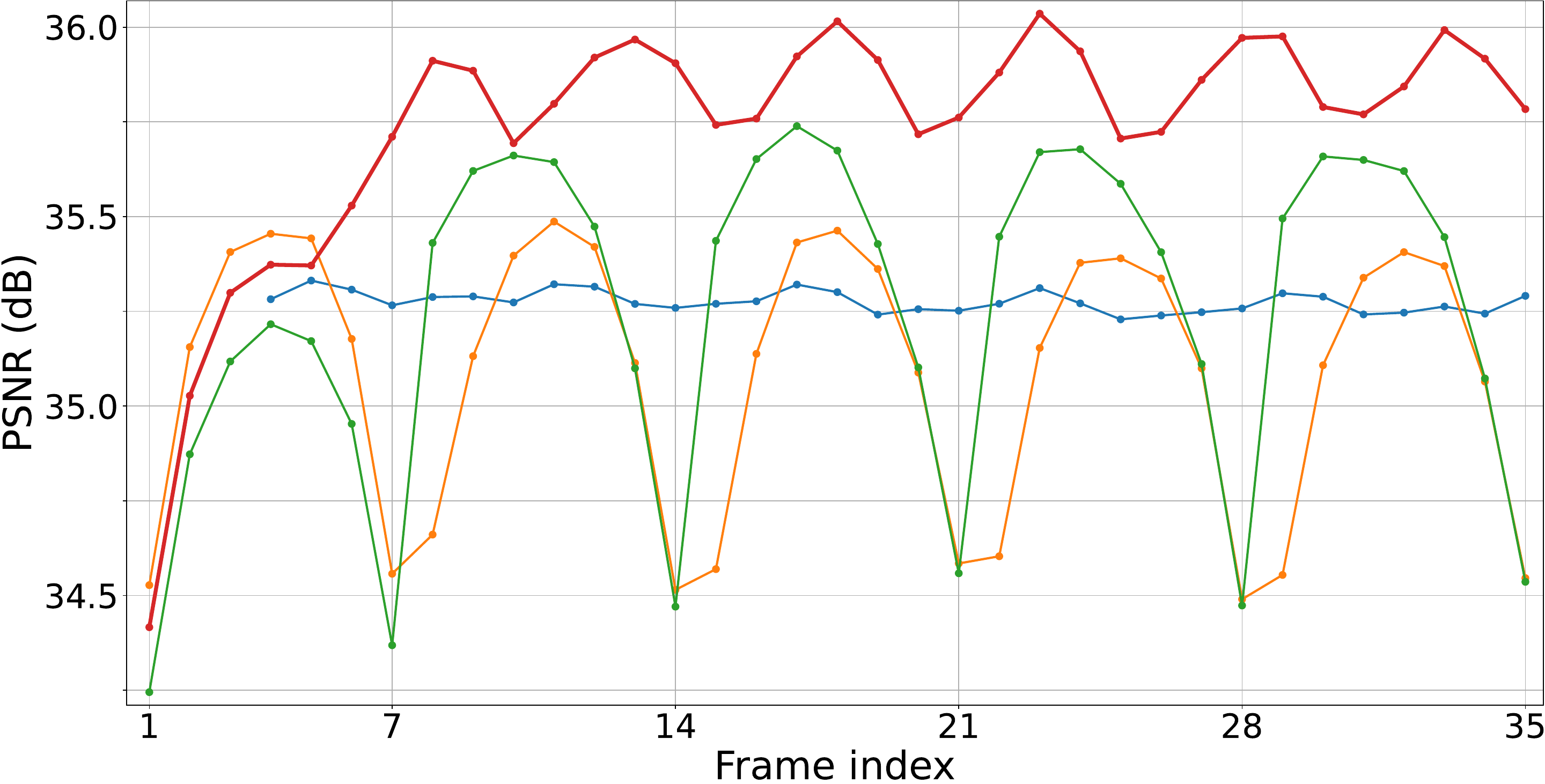} & \includegraphics[width=0.47\linewidth]{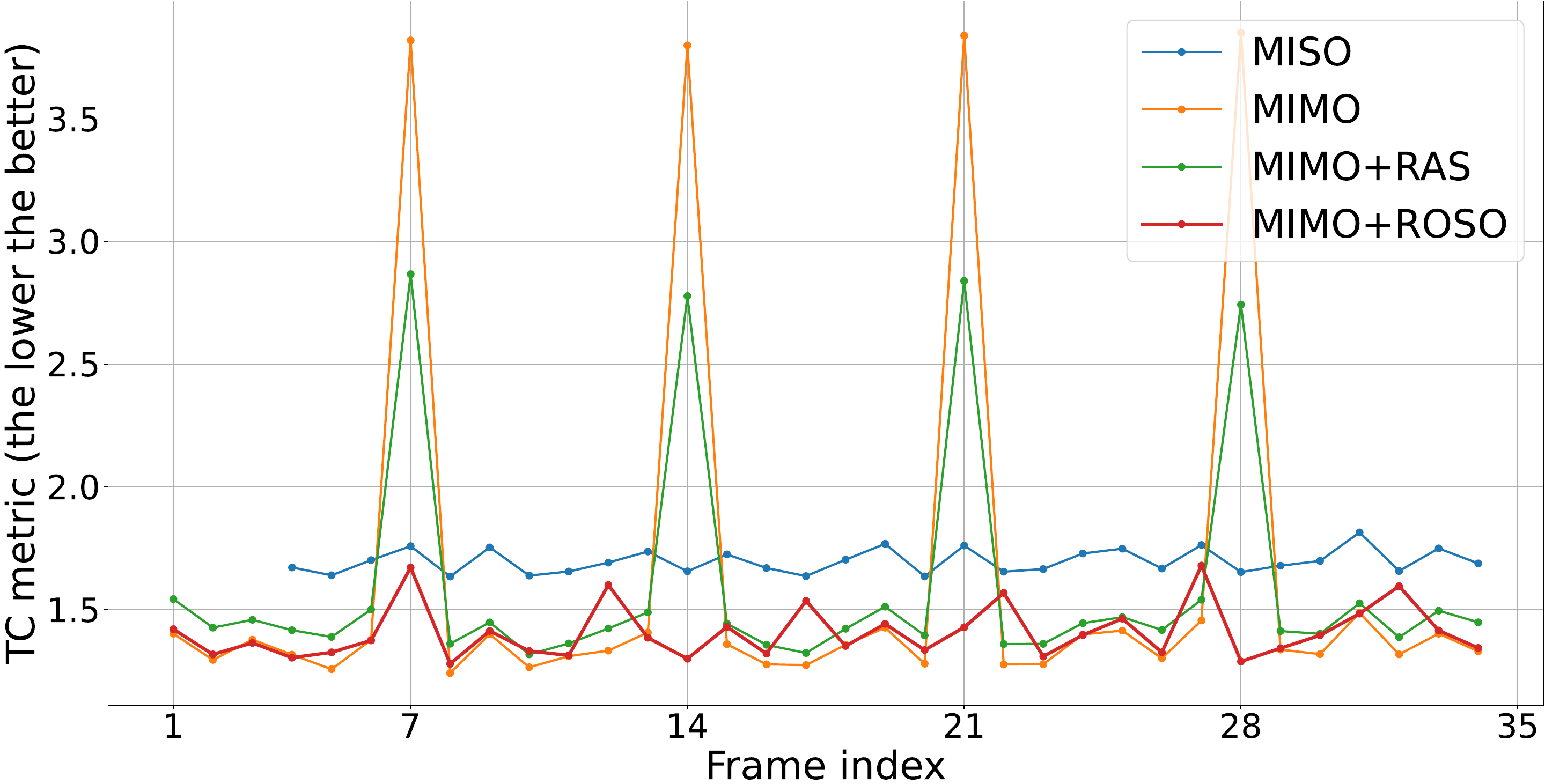}\\(a) PSNR & (b) Time consistency between frames 
    \end{tabular}
    \vspace{-.5em}
    
  \caption{(a) The PSNR of MIMO networks is not uniform within the output stack. We propose to use recurrence across stacks (RAS) and output stack overlap (OSO) to smooth this non uniformity.
(b) MIMO networks are temporally consistent within output stacks but have strong changes at stack transitions. The proposed contributions (shortened to ROSO when applied together) significantly reduce these changes.
The reported plots show for each frame, the average PSNR over the drone benchmark, for denoising of AWGN $\sigma=
30$ with the M2Mnet network~\cite{chen2021multiframe}.}
    \label{fig:teaser}
\end{figure}

\section{Introduction}

The state of the art in video restoration is currently dominated by neural networks trained with supervision on large datasets. 
%
Early video restoration architectures~\cite{vnlnet_icip2019,tassano2020fastdvdnet} exploit the temporal redundancy of videos by taking multiple input frames (that we denote a stack) to produce a single output frame (MISO) for each network evaluation.
More recently, the dominating architectural trend \cite{xiang2022remonet,chen2021multiframe,lindner2023lightweight,cao2023learning,chan2022basicvsr++,Li_2023_CVPR,liang2024vrt}, is based on multiple-input, multiple-output (MIMO) networks, which produce several output frames per network evaluation and process the video in non-overlapping stacks of frames. This improves computational efficiency by sharing computations across multiple frames, thus reducing the per-frame computational cost.  In addition, output frames within the same stack tend to have a much higher temporal consistency.

The top performing networks in video  restoration~\cite{chan2022basicvsr++,Li_2023_CVPR,liang2024vrt,liang2022recurrent}, 
take this idea to the extreme by processing all the frames that fit in the GPU memory in each network evaluation. This has led to results of unprecedented quality, at the cost of high computational cost and latency, resulting in methods that are not suitable for real-time, low-latency applications.

%

In this work we focus on the application of MIMO networks to the low-latency scenario, with latency defined as the maximum number of future frames needed to produce each output frame. This restricts the size of the MIMO stack, limiting in turn the achievable quality, as it heavily depends on the temporal receptive field of the network. Frames near the borders of the MIMO stack are more affected by this, which leads to a characteristic rectified sine profile of the per-frame PSNR plot \cite{chan2021basicvsr}.
An example is shown in Figure~\ref{fig:teaser}(a).

In addition, the strong temporal consistency within the MIMO stack contrasts with the temporal discontinuity at stack transitions (see Figure~\ref{fig:teaser}(b)), inducing a periodic step-wise motion artifact.
This effect is present regardless of the size of the MIMO stack, but for small stack sizes it becomes especially conspicuous. 
{To the best of our knowledge, this step-wise motion artifact has not been reported before, probably in part due to its masking, as existing benchmarks consist of hand-held videos with severe camera shake~\cite{khoreva2019video}.}
However, as soon as these videos are stabilized the periodic temporal inconsistencies become noticeable again.


Our contributions in this work are the following:

\smallskip \noindent \textit{(i)} We provide an empirical analysis of MIMO networks in a low-latency setting, demonstrating two significant problems: the quality decay of the restored frames at the boundary of the output stacks; and the step-wise motion artifact induced by stack transitions. 
While the first issue has been described 
{before}~\cite{qi2022real,chan2021basicvsr}, we could not find any mention of the second. 

\smallskip \noindent \textit{(ii)} 
We propose to address these issues by two strategies: recurrence \emph{across stacks} (RAS) and output stack overlap (OSO). 
Recurrence across stacks improves the output quality by implicitly enlarging the temporal receptive field, specially for the first frames of the stack; while output stack overlap allows to remove almost completely temporal inconsistencies and  further improves the output quality (at a $40\%$ to $60\%$ increase in running).  These strategies can be applied to any MIMO video restoration architecture. We demonstrate their effectiveness on three state-of-the-art networks with different architectures. As shown in the left curve of \Cref{tab:results_on_drone_dataset}, the proposed strategies improve over the current Pareto frontier on the PSNR-vs-runtime landscape among low-latency networks.

\smallskip \noindent \textit{(iii)} 
As a complement to the existing benchmarks, we introduce a new 
evaluation dataset of 14 stabilized videos taken with drone mounted cameras. This dataset has a very smooth motion 
which highlights temporal consistency issues that are masked on existing benchmarks. 
As most videos are stabilized before human consumption, we argue that stabilized video is a relevant use-case for video restoration, which is under-represented in the standard benchmarks. 
%
\smallskip

Throughout the paper we focus on the problem of denoising RGB video with additive white Gaussian noise (AWGN).
This is a rather artificial problem, but the findings of our work should apply to MIMO networks for similar video restoration problems (such as super-resolution and deblurring), particularly if the input data is contaminated with noise as is usually the case in practical applications. 




\section{Related work}

\myparagraph{MIMO architectures.} 
Since their introduction, MIMO networks have been widely used for multiple video restoration tasks~\cite{chan2022basicvsr++,xiang2022remonet,chen2021multiframe,Li_2023_CVPR,liang2024vrt} (denoising, deblurring, super-resolution) especially because they decrease the per frame processing time.
It has been pointed out that some MIMO networks suffer from a performance drop near the stack transitions, translating in a distinctive periodic pattern in a per frame PSNR plot~\cite{chan2021basicvsr}.
Recurrence has been suggested as a way to attenuate this effect~\cite{xiang2022remonet}.

\myparagraph{Recurrent networks.} 
Recurrent architectures allow the network to have a \emph{memory} of previous past frames, while still evaluating on a small input window. 
Recurrent MISO networks have been proposed for video denoising~\cite{maggioni2021efficient,huang2022neural,ostrowski2022bp,chan2022generalization}, 
super-resolution~\cite{yang2018video,sajjadi2018frame,isobe2020video,fuoli2019efficient} and video joint denoising and demosaicing~\cite{dewil2023video}.
MIMO networks such as BasicVSR++~\cite{chan2021basicvsr,chan2022basicvsr++} and RVRT~\cite{liang2022recurrent} use recurrence as a way of propagating temporal information within stacks. ReMoNet~\cite{xiang2022remonet} is a video denoising network which uses recurrence \emph{across} stacks. 
In this work we reproduce the results of ReMoNet and demonstrate the effectiveness of the approach with two other MIMO architectures. Our experiments show that across-stack recurrence contributes to increasing the video quality but does not eliminate the temporal discontinuities at stack transitions.

\myparagraph{Temporal consistency.} 
Temporal consistency is usually enforced by incorporating regularization terms. 
Some methods use terms that enforce consistency under simulated motion~\cite{eilertsen2019single,zheng2016improving}. Penalizing the difference between a recovered frame and the warped version of a previous recovered frame is used for colorization~\cite{lai2018learning,lei2019fully,Zhang_2019_CVPR,zhao2022vcgan,liu2024temporally}, style transfer~\cite{gupta2017characterizing,gao2019reconet}, and also in classic energy minimization methods for  inpainting~\cite{kokaram2005automated,shiratori2006video,bhat2007using,bhat2010gradientshop} or video editing ~\cite{facciolo2011temporally,sadek2013variational}.
Incorporating such a regularization term in the training loss has been proposed for MIMO video denoising~\cite{chen2021multiframe}. In contrast, our proposed OSO approach does not require to complexify the training loss while being consistent at stack transitions.

\section{Proposed method}\label{sec:method}

A MIMO network is applied to stacks of $N_i$ input frames and produces $N_o$ output frames. Typically $N_i = N_o + 2N_p$, where $N_p$ is a number of padding frames added at both ends of the stack. As discussed earlier, MIMO networks provide a very good trade-off between output quality and running time. 
In this work we focus on applications that require low-latency. We define latency as the maximum number of future frames needed to produce an output frame%
{,  ignoring the network evaluation time.} For a network with $N_o$ frames and $N_p$ input padding frames the worse case latency is $N_o + N_p - 1$.  
{Several video restoration networks proposed in the literature use small stacks with $5$ to $7$ frames~\cite{xiang2022remonet,tassano2019dvdnet,tassano2020fastdvdnet,sheth2021unsupervised}. Other networks allow a variable stack size, and use as many frames as they can fit in the GPU memory, as the output quality increases with the stack size~\cite{Li_2023_CVPR,chan2022generalization,liang2024vrt,liang2022recurrent}.}

In this work we limit the stack size to 7 frames (exceptionally 8 for architectures that require even stack sizes) as we are interested in low-latency applications. In the next sections we describe two problems of MIMO networks that are aggravated for small stack sizes, and propose solutions to them.


\subsection{Output stack overlapping (OSO)} 
\label{subsec:oso}

Temporal consistency is a key aspect of perceived video quality.
Within an output stack, 
{MIMO networks have a higher temporal consistency than MISO.}
However, at stack transitions
MIMO networks have strong temporal discontinuities
that contrast with the temporal smoothness within the stacks. 
This is highlighted in~\Cref{fig:teaser}(b), where we measure the temporal consistency for each frame as the temporal $L_1$ warping error, i.e. as the mean absolute difference between one frame and the aligned version (using an optical-flow~\cite{zach2007duality,Perez2013}) of the next one. 
See also \Cref{fig:results_at_stack_transition} for a visual example.
In the low-latency setting this is perceived as a step-wise motion, as if the framerate was lower by a factor of $1/N_o$. This effect is especially notorious in areas with low signal-to-noise ratio (SNR), such as textures with a contrast lower than the noise level (see videos in supp. mat.).




To prevent these discontinuities, we propose to apply the network to overlapping stacks, so that the last $P$ frames of an output stack overlap with the first $P$ frames of the next one. Each frame at the overlap region is therefore estimated twice. We denote these estimates 
{$\hat u^s_i$ and $\hat u^{s+1}_i$}
, where $i = 1,...,P$ denotes the index in the overlap region and $s$ is the stack number. 
This is illustrated in \Cref{fig:OSO_framework}(b).
To smooth the stack transition, the two estimates of each frame are combined using a weighted average 
$\hat u_i = (1-\alpha_i) \hat u^s_i + \alpha_i\hat u^{s+1}_i$ 
where $\alpha_i = i/(P+1)$. 
Thus $\hat u_i$ transitions linearly from stack $s$ to $s+1$. 
Compared to the non-overlapping framework, this increases the computational complexity by a factor of $N_o /(N_o-P)$.



\subsection{Recurrence across stacks (RAS)}  

{Limiting the latency of MIMO networks 
amounts to restricting the size of the input stack $N_i$, which in turn limits the output quality.}
Furthermore, the performance of MIMO networks is not uniform inside the output stack as output frames at the borders of the stack have a lower quality than the frames towards the center. This leads to a characteristic {rectified sine} PSNR profile (\Cref{fig:teaser}(a)). 
For large MIMO stacks, this affects a small fraction of the frames. 
However, in the low-latency setting, this profile is repeated every $N_o$ frames (usually $N_o=5$ or $7$), thus several times per second. 

\begin{figure}
    \centering
    \includegraphics[width=\textwidth]{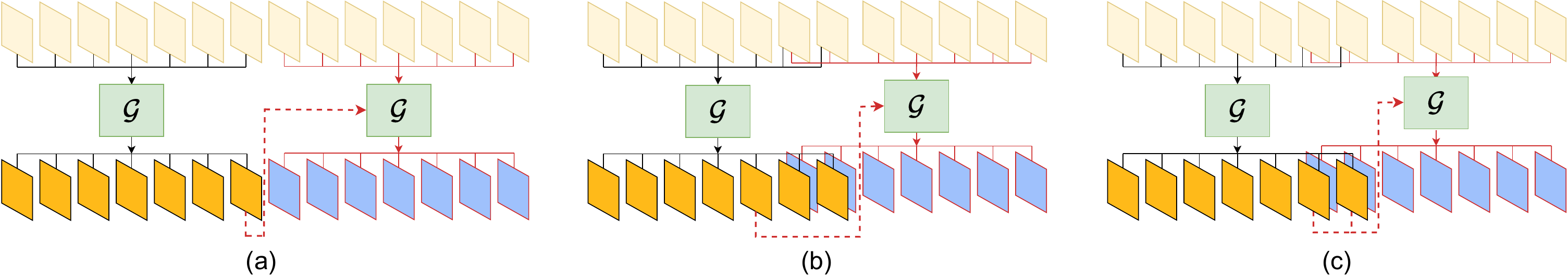}
    \vspace{-2em}
    \caption{MIMO and OSO architectures. 
    {(a) Baseline MIMO network applied to two non-overlapping frame stacks (b) Output stack overlap (OSO). The output in the overlapped frames is computed as a weighted average of the two denoised versions. Dashed lines in (a) and (b) indicate recurrence across stack (RAS) (c) OSO with RAS with a specific overlapped recurrence using the output overlapped frames from the previous stack.}
    }   
    \label{fig:OSO_framework}
\end{figure}

One natural way of implicitly enlarging the temporal receptive field of a network is by using recurrence.
Some MIMO networks use recurrent layers as a mechanism for temporal propagation \emph{within} the stack. But here we are interested in having recurrence \emph{across stacks.}
This can be expressed as follows
\begin{equation}
(\hat u^s_{t_s-\lfloor N_\mathrm{o}/2\rfloor},...,\hat u^s_{t_s+\lfloor N_\mathrm{o}/2\rfloor}, m^s) = \mathcal G(m^{s-1},\,\, f_{t_s-\lfloor N_\mathrm{i}/2 \rfloor},...,f_{t_s+\lfloor N_\mathrm{i}/2 \rfloor}),
\end{equation}
where $\mathcal{G}$ denotes the network, $s$ the stack number, $\hat u^s_t$ denotes the output frames, $f_t$ the input frames and $m^{s-1}$ acts as a memory which is updated with each network evaluation and passed to the next one.
Several choices for 
$m^{s-1}$ have been proposed such as a feature map computed by an intermediate network~\cite{sajjadi2018frame,xiang2022remonet}, or some of the output frames from the previous stack.
The output stack overlapping provides a new interesting choice for $m^{s-1}$, namely using the $P$ frames in the overlap region 
{$\hat u^{s-1}_{t_{s-1}+\lfloor N_\mathrm{o}/2\rfloor-P+1}, \dots,\hat u^{s-1}_{t_{s-1}+\lfloor N_\mathrm{o}/2\rfloor}$}, as shown in~\Cref{fig:OSO_framework}(c).
We refer to this as \emph{overlapped output recurrence}. 
The network at stack $s$ thus has a direct preliminary estimate of the first $P$ frames of its output stack. 

We also consider for $m^{s-1}$ the output frame of stack $s-1$ preceding the first output frame of stack $s$, i.e. 
{$\hat u^{s-1}_{t_{s-1}+\lfloor N_\mathrm{o}/2\rfloor - P}$}. This option can be used either with OSO, in which case it is the last output of stack $s-1$ before the overlap region; or without OSO, in which case it is the last output frame of stack $s-1$. In both cases, this frame is not an estimate of any of the outputs of the current stack $s$, thus it will be miss-aligned.
We refer to this as \emph{previous output recurrence}.
The mechanics of these different options are illustrated in~\Cref{fig:OSO_framework}(a-b), with and without OSO.

As shown in~\Cref{fig:teaser}(b), adding recurrence across stacks already mitigates the temporal inconsistency at stack transitions. Although this reduction is small, the use of recurrence significantly increases output quality. 


\subsection{Networks}

We selected three MIMO networks for our experiments, covering different ranges of the Pareto frontier depicted in \Cref{tab:results_on_drone_dataset}. 
{In the following, we give a brief description of each.}
More details can be found in the supplementary material.

\myparagraph{BasicVSR++.} Initially introduced for video super-resolution~\cite{chan2022basicvsr++}, it has been  later extended to video denoising and deblurring~\cite{chan2022generalization}. 
It uses deformable convolutions~\cite{dai2017deformable}, guided by an optical flow~\cite{ranjan2017optical} to aligned features extracted from temporal input frames. 
It works with large MIMO stacks determined by the GPU memory (referred as \emph{BasicVSR++/full} in Section~\ref{sec:results}). 
We adapt it to the low-latency case by simply applying it on non-overlapping stacks of size $N_i=7$ frames. We also made a variant with recurrence across stack by modifying the code from the authors to propagate the features of the previous stack into the next.
In all variants, we use the weights provided by the authors.

\myparagraph{M2Mnet.} M2Mnet~\cite{chen2021multiframe} takes as input $N_i=7$ frames and outputs $N_o=7$ denoised frames. 
{We modified the version described in~\cite{chen2021multiframe}  (baseline) and  made it recurrent.}

\myparagraph{ReMoNet.} 
To the best of our knowledge, ReMoNet~\cite{xiang2022remonet} is the only MIMO network in the video restoration literature with recurrence across stacks.
ReMoNet~\cite{xiang2022remonet} follows a similar structure to FastDVDnet~\cite{tassano2020fastdvdnet}. It consists of two cascaded U-Nets considered as two denoising stages. The first U-Net is recurrent (\emph{previous} recurrence). To apply RAS, we reuse the last recurrence map when we process the next stack.
We use $N_i=7, N_o = 5$ frames.
\smallskip

As there are no public implementations available for M2Mnet and ReMoNet, we implemented and trained these architectures  according to the information provided in the papers. More details about the architectures are given in the supplementary material. 
For the recurrent networks, we initialized the recurrent frame with zero-value feature maps.

\section{Results}\label{sec:results}

We test the impact of the proposed improvements on MIMO networks in the task of denoising RGB videos corrupted with additive white Gaussian noise.
Most video denoising methods are benchmarked on the DAVIS-17~\cite{Pont-Tuset_arXiv_2017}, REDS4~\cite{Nah_2019_CVPR_Workshops_REDS} or Set8~\cite{tassano2020fastdvdnet}. 
These validation sets have been downsampled temporally \cite{Pont-Tuset_arXiv_2017} or acquired using a hand-held camera \cite{Nah_2019_CVPR_Workshops_REDS} resulting in strong motions and camera shake.
We argue that it is also necessary to evaluate video restoration performance on videos with stabilized motion. Indeed such videos highlight problems with temporal consistency that are masked in videos with fast and irregular motion.
To that aim we introduce a validation dataset consisting of 14 sequences of 40 960$\times$540 frames each extracted from videos of drones.\footnote{Available at \url{http://xxx.yyy.zzz}}
We will report results in the proposed drone benchmark, as well as in the standard benchmarks.

\myparagraph{Networks and training.} Our experiments are done with M2Mnet~\cite{chen2021multiframe}, ReMoNet~\cite{xiang2022remonet} and BasicVSR++~\cite{chan2022generalization}. For each, $N_i=7$ and $N_o=7$ except for ReMoNet, which discards 2 frames so $N_o=5$ in its case. We apply OSO by overlapping $P=2$ frames. 
For BasicVSR++ we use the available weights. For ReMoNet and M2Mnet we use our own implementation trained from scratch.
We use two training sets: a drone-specific training set consisting of drone videos taken from 
UAVs~\cite{li2016multi}, ERA~\cite{eradataset} and Visdrone~\cite{9573394}; and a generic training set formed by DeepVideoDeblurring~\cite{su2017deep} and train-REDS~\cite{Nah_2019_CVPR_Workshops_REDS} with and without stabilization (as in~\cite{dewil2023video}).

\begin{table}[t!]
\center
\begin{minipage}{0.55\linewidth}
\includegraphics[width=\linewidth]{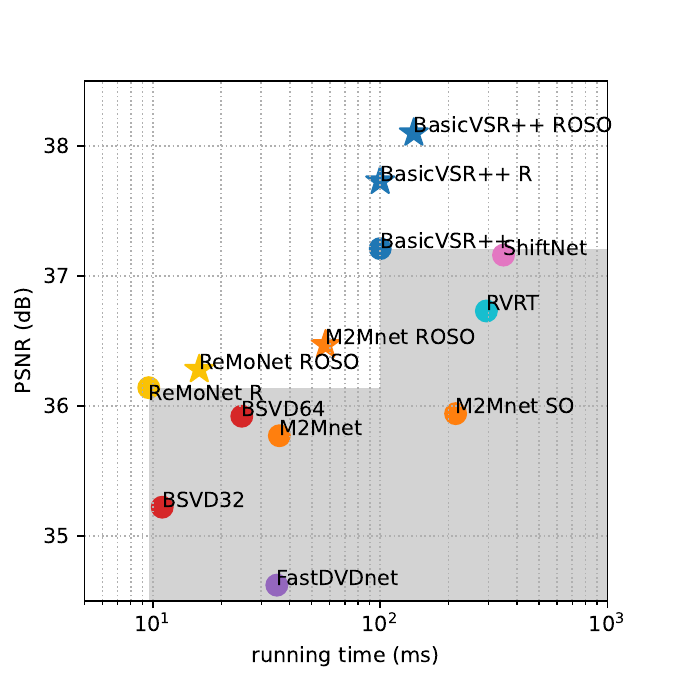}
\end{minipage}%
\begin{minipage}{0.45\linewidth}
\scalebox{0.74}{
\begin{tabular}{ %
|c|c|c|c|c|c|}
    \hline
    & \rotatebox[origin=c]{90}{MIMO} & \rotatebox[origin=c]{90}{RAS} & \rotatebox[origin=c]{90}{OSO} & PSNR / SSIM & TC intra/inter \\ \hline
    \multirow{7}{.5em}{\rotatebox[origin=c]{90}{M2Mnet}} 
    &            &   &            & 35.94 / 0.957 & 1.811 / 1.937\\
    &            & P &            & 35.85 / 0.956 & 1.704 / 1.800\\
    & \checkmark &   &            & 35.77 / 0.956 & 1.445 / 3.727\\
    & \checkmark & P &            & 35.97 / 0.960 & 1.537 / 2.672\\ \cdashline{2-6}
    & \checkmark &   & \checkmark & 36.11 / 0.959 & 1.519 / 1.769\\
    & \checkmark & P & \checkmark & \textbf{36.49} / \textbf{0.963} & 1.467 / \textbf{1.671}\\
    & \checkmark & O & \checkmark & 36.47 / \textbf{0.963}&\textbf{1.457} / 1.689\\ \hline \hline                             
    \multirow{7}{.5em}{\rotatebox[origin=c]{90}{ReMoNet}}
    &            &   &            & 35.77 / 0.958 & 1.773 / 1.884\\
    &            & P &            & 36.14 / 0.962 &1.590 / 1.712\\
    & \checkmark &   &            & 35.39 / 0.954 &1.522 / 2.928\\
    & \checkmark & P &            & 36.14 / 0.962 & \textbf{1.467} / 2.382\\ \cdashline{2-6}
    & \checkmark &   & \checkmark & 35.56 / 0.956 &1.628 / 1.821 \\
    & \checkmark & P & \checkmark & \textbf{36.28} / \textbf{0.963} &1.528 / \textbf{1.678} \\ \hline \hline
    \multirow{3}{.5em}{\rotatebox[origin=c]{90}{{\scriptsize BasicVSR++}}}  
    & \checkmark &   &            & 37.21 / 0.966 & \textbf{1.173} / 2.886\\
    & \checkmark & P &            & 37.73 / 0.971 & 1.212 / 2.602 \\\cdashline{2-6}
    & \checkmark & P & \checkmark & \textbf{38.10} / \textbf{0.973} & 1.278 / \textbf{1.529}\\ \hline
\end{tabular}}
\end{minipage}

\caption{Right: Average PSNR/SSIM/TC metrics of various MIMO architectures for Gaussian noise ($\sigma=10,20,30,40,50$) on our val-drone set. 
RAS uses either the previous frame (P) or the overlapped frames (O).
Left: Landscape of {low-latency} video denoising networks derived from the results presented in the papers. The vertical axis shows the PSNR obtained on the drone benchmark, averaged over noise levels $\sigma=10, 20, 30, 40, 50$. The methods plotted with a star are variants proposed in this paper. 
}

\label{tab:results_on_drone_dataset}
\end{table}

\myparagraph{Temporal consistency metrics.} 
From the per-frame temporal $L_1$ warping error defined in Section~\ref{subsec:oso}, we derive two single-value metrics 
that summarize the temporal consistency of a denoised video. 
The \emph{intra-stack} TC
is computed as the median of the per-frame TC values in the sequence, and will therefore filter out the spikes at stack boundaries. 
The \emph{inter-stack} TC is computed  
by first taking the maxima of the per-frame TC over non-overlapping windows of $N_o$ frames (thus selecting the spikes at stack transitions), and then taking the median of these maxima. 
Temporal consistency metrics based on warping error have been used before as quality metrics or loss functions (\textit{e.g.} \cite{kokaram2005automated,gupta2017characterizing}). They rely on the brightness constancy assumption \cite{horn1981determining} and thus are no suitable for videos with illumination changes, reflections and semi-transparency. Gradient-based metrics robust to illumination changes have also been proposed \cite{bhat2010gradientshop,sadek2013variational}.

{For the following numerical results, we report for each dataset and network the average PSNR, SSIM, and TC metrics computed across sequences and the five noise levels ($\sigma=10,20,30,40$ and $50$). For each network, the averages per noise level can be found in the supplementary material.}

\begin{table}
\center
\scalebox{0.85}{
\begin{tabular}{p{0.2cm} |p{2.4cm}||p{1.0cm} |p{1.5cm}|p{1.5cm} |p{1.5cm}|p{1.5cm}||p{0.8cm}|p{0.8cm}|}
    \cline{1-9} 
    & \multirow{2}{4em}{Network}  &\multirow{2}{4em}{r.t. (ms)} & \multirow{2}{4em}{DAVIS} & \multirow{2}{4em}{Set8} &  \multirow{2}{4em}{REDS4} & REDS4 & \multicolumn{2}{c|}{T.C.}\\ 
    & & &  & & & stabilized & \emph{intra} & \emph{inter}\\ \cline{1-9}
    \multirow{11}{1em}{{\rotatebox[origin=c]{90}{low latency}}}
    &M2Mnet    &  \hfill 36.0 & 35.11/.927 & 32.61/.895 & 33.77/.908  & 34.80/.930 & 2.889 & 4.820 \\
    &M2Mnet ROSO        &  \hfill 57.5 & \textbf{35.30/.929} & \textbf{32.84/.901} & \textbf{33.85/.910} & \textbf{34.98/.933} & \textbf{2.878}  & \textbf{3.288}\\ \cline{2-9}
    &ReMoNet      & \hfill 9.6 &34.66/.920 &32.23/.890 &33.41/.902 &34.25/.923 &3.600 &4.289 \\
    &ReMoNet ROSO          & \hfill 16.0 & \textbf{34.74/.921} &\textbf{32.29/.892} &\textbf{33.45/.903} &\textbf{34.33/.924} &\textbf{3.481} &\textbf{3.770} \\ \cline{2-9} 
    &B.VSR++  & \hfill 100.1 & 36.40/.943 & 33.79/.914 & 35.73/.939 & 36.10/.948 & \textbf{2.036} & 4.002\\   
    &B.VSR++ ROSO       & \hfill 140.1  & \textbf{37.00/.951} & \textbf{34.33/.925} & \textbf{36.33/.947} & \textbf{36.71/.955} & 2.067 & \textbf{2.445} \\ \cline{2-9}   
    &Shift-net (plus)   & \hfill 348.8 & 36.39/.952 & 33.74/.913 & 35.21/.932 & 36.04/.947 & 2.061 & 4.112 \\ 
                   
    \cline{2-9} 
                    
    &RVRT   & \hfill 293.3 & 36.6/.938 & 35.04/.913 & 35.94/.932 & 36.4/.945 & 2.355 & 3.993 \\                    
    \cline{2-9} 
    &FastDVDnet& \hfill 35.1 &  34.79/.921 & 32.46/.890 & 33.40/.901 & 34.29/.923 & 3.475 & 3.744\\
     \hline \hline
    \multirow{4}{1em}{{\rotatebox[origin=c]{90}{full}}}

    &BasicVSR++   & \hfill 100.1 & 37.49/.957 & 34.86/.934 & 36.72/.952 & 37.12/.960 & 2.024 & 2.350\\ \cline{2-9} 
    &RVRT      & \hfill 293.3 & 37.21/.948 & 35.69/.929 & 36.46/.942 & 37.04/0.955 & 2.288 & 2.604 \\ \cline{2-9} 
    &Shift-net (plus) & \hfill 348.8 & 37.08/.952 & 34.31/.925 & 35.64/.938 & 36.70/.955 & 2.044 & 2.571 \\ 
    \cline{1-9} 
\end{tabular}}
\caption{Average PSNR/SSIM/TC metrics for Gaussian noise ($\sigma=10,20,30,40,50$) for the three standard benchmark video datasets. The two single-value temporal consistency metrics are computed on the REDS stabilized dataset. For each low-latency network, best performance is in bold. The networks at the top are low-latency, and their full video versions are at the bottom where applicable.}
\label{tab:benchmarks_table}
\end{table}

\myparagraph{Ablation study.} 
Table~\ref{tab:results_on_drone_dataset} (right) shows an ablation study of M2Mnet, ReMoNet and BasicVSR++ on the drone benchmark. We compare the performances of these architectures with different configurations MISO or MIMO, with/without RAS and with/without OSO.

\smallskip\noindent{\it MISO vs. MIMO.} MISO versions of M2MNet and ReMoNet (non-recurrent) attain higher PSNR and SSIM than the MIMO counterparts. This is to be expected as the MIMO task is more complex. Note also that temporal consistency metrics of MISO networks lie between the intra and inter TC of the MIMO versions (as seen also in Figure~\ref{fig:teaser}).

\smallskip\noindent{\it RAS.} With recurrence across stacks, MIMO networks meet (even surpass) the MISO performance. This is largely due to improving the first few frames in each output stack (Figure~\ref{fig:teaser}).
Recurrence also improves the results of BasicVSR++. 
For all methods, RAS reduces the temporal discontinuity at stack transitions (Figure~\ref{fig:teaser}(b)) leading to a lower (yet still perceivable) inter TC.


\smallskip\noindent{\it OSO.} 
For the three MIMO networks, either with or without RAS, OSO drastically reduces inter stack TC (at the expense of a slight increase in intra stack TC), and also improves PSNR and SSIM.
Best results are obtained with combining recurrence and OSO, which we will denote by ROSO.

\smallskip\noindent{\it Overlapped vs previous output RAS.}  In the metrics averaged across noise levels for the drone benchmark shown in Table~\ref{tab:results_on_drone_dataset} (right), there is no difference between \emph{overlapped} and \emph{previous frame} RAS for M2Mnet.
\emph{Overlapped} RAS works better for noise $\sigma \ge 30$ and in the standard benchmarks for all noise levels (see supplementary). 
This is likely due to the fact that with stronger motion it becomes difficult for the network to use the previous output frame as it is not aligned with the current outputs. 

\myparagraph{The PSNR-time landscape.}
The plot in Table~\ref{tab:results_on_drone_dataset} (left) shows several video denoising networks mapped in a PSNR vs per-frame running time landscape computed on the drone benchmark. The running times were measured on a Nvidia A100 GPU. Some of them are reported in Table~\ref{tab:benchmarks_table}, and the rest can be found in the supplementary material.
For methods that can use a large number of frames, we used their low-latency variant
with a limited window size of $N_i = 7$ (i.e., latency of 7 frames). This provides a relevant comparison for  restricted latency settings. In addition, by using all architectures with similar number of input frames, it shows a more meaningful comparison of their performance in equal conditions.

Using this visualization, we can identify a Pareto frontier showing the optimal trade-off between computational cost and PSNR. 
Methods in the gray region are sub-optimal in the sense that another it is possible to achieve higher PSNR at lower or equal computational cost.
With the exception of BSVD~\cite{qi2022real}\footnote{BSVD~\cite{qi2022real} produces a single output frame per network evaluation, but it avoids repeating redundant computations for nearby frames, by storing the outputs of intermediate layers into buffers that are shared between network evaluations.}
all networks close to the Pareto frontier are MIMO, namely BasicVSR++~\cite{chan2022basicvsr++}, Shift-Net~\cite{Li_2023_CVPR} and our own implementation of M2Mnet~\cite{chen2021multiframe}. 
The proposed ROSO variants define a new Pareto frontier due to the improved PSNR, in spite of the increase in the running time (between 40\% and 60\% longer, which is consistent with expected increase of $N_o/(N_o - P) = 7/5$).

\myparagraph{Standard evaluation benchmarks.} In Table~\ref{tab:benchmarks_table}, we compare the tested networks  (\emph{baseline} and ROSO) with other state-of-the-art video denoising networks on the standard RGB video denoising benchmarks.

Table~\ref{tab:benchmarks_table} shows that the state-of-the-art methods in their \emph{full} configuration (BasicVSR++, Shift-net, and RVRT) 
provide the best PSNR, SSIM and TC metrics. This is to be expected due to the 
larger temporal receptive field. However, when we compare the low-latency versions of these MIMO networks (denoising independent stacks of 7 to 8 frames), we observe that their respective performance drops significantly. 
Comparing the ROSO versions of M2MNet, ReMoNet and BasicVSR++ with their respective baselines we conclude that the ROSO versions always have superior performance, albeit with a smaller margin than in the case of the drone dataset. The proposed improvements have a larger impact on videos with smooth and stable motion, as those obtains with motion stabilization and high frame rates.

Lastly we note that the results of all networks except BasicVSR++ present a large performance gap (of about 1dB) between the REDS4 dataset and its stabilized version. 
This indicates that stabilization is key for simple architectures as M2MNet and ReMoNet.  BasicVSR++ performs similarly regardless of the stabilization probably due its motion compensation mechanism.

\begin{figure}[t!]
\center
\setlength{\tabcolsep}{1pt} 
\def\s{0.19\textwidth}

\setlength{\tabcolsep}{1pt} 
\small
\def\seqone{d160-027}
\def\seqtwo{d160-028}
\def\seqthree{d160-029}
\begin{tabular}{ccccc}
     stack $N$; $t-2$ & & stack $N$; $t-1$ & $\leftarrow$\textbf{transition}$\rightarrow$  & stack $N+1$; $t$ \\
    \overimg[width=\s,trim={80 0 80 30}, clip]{figures/visual_results/gt-drone\seqone.png}{GT} & \includegraphics[width=\s,trim={80 0 80 30}, clip]{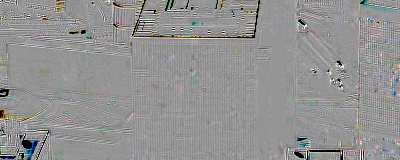} &\includegraphics[width=\s,trim={80 0 80 30}, clip]{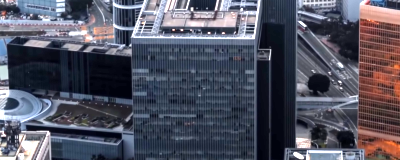} & \includegraphics[width=\s,trim={80 0 80 30}, clip]{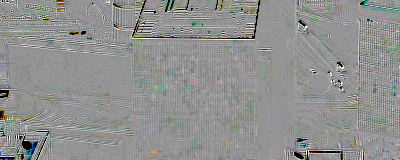} & \includegraphics[width=\s,trim={80 0 80 30}, clip]{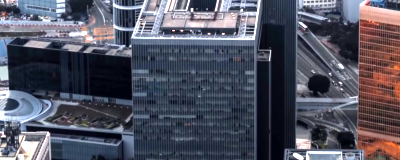} \\
    
     \overimg[width=\s,trim={80 0 80 30}, clip]{figures/visual_results/noisy-drone\seqone.png}{noisy} & \includegraphics[width=\s,trim={80 0 80 30}, clip]{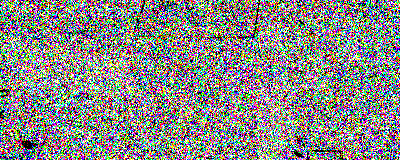} & \includegraphics[width=\s,trim={80 0 80 30}, clip]{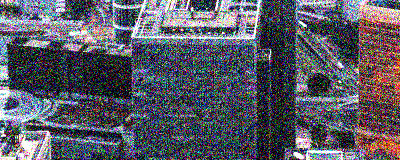} & \includegraphics[width=\s,trim={80 0 80 30}, clip]{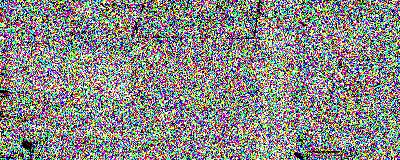} & \includegraphics[width=\s,trim={80 0 80 30}, clip]{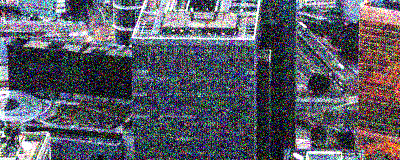}  \\
    
     \overimg[width=\s,trim={80 0 80 30}, clip]{figures/visual_results/M2Mnet_baseline-drone\seqone.png}{baseline} & \includegraphics[width=\s,trim={80 0 80 30}, clip]{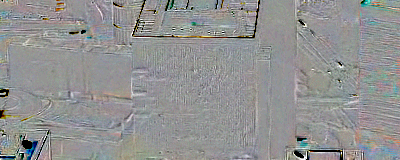} & \includegraphics[width=\s,trim={80 0 80 30}, clip]{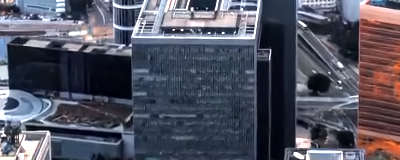} & \includegraphics[width=\s,trim={80 0 80 30}, clip]{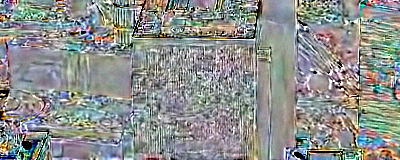} & \includegraphics[width=\s,trim={80 0 80 30}, clip]{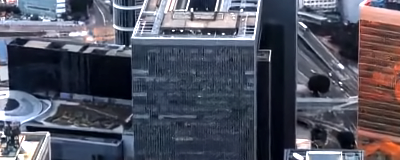}  \\

    \overimg[width=\s,trim={80 0 80 30}, clip]{figures/visual_results/M2Mnet_baselineRec-drone\seqone.png}{RAS} & \includegraphics[width=\s,trim={80 0 80 30}, clip]{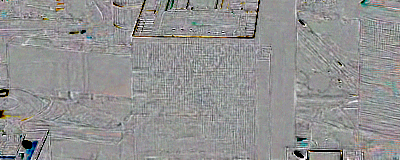} & \includegraphics[width=\s,trim={80 0 80 30}, clip]{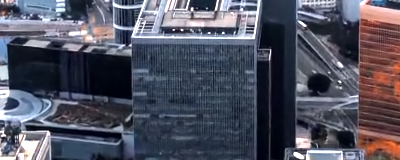} & \includegraphics[width=\s,trim={80 0 80 30}, clip]{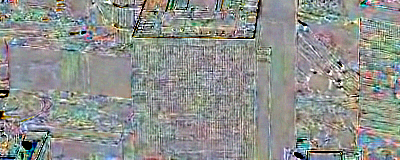} & \includegraphics[width=\s,trim={80 0 80 30}, clip]{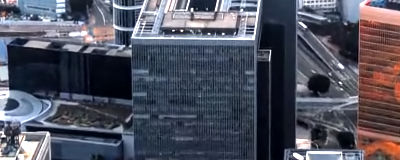} \\

    \overimg[width=\s,trim={80 0 80 30}, clip]{figures/visual_results/M2Mnet_baselineRecOSO-drone\seqone.png}{ROSO} & \includegraphics[width=\s,trim={80 0 80 30}, clip]{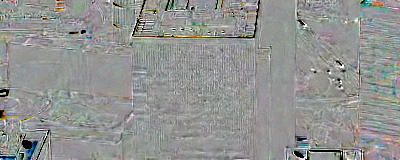} & \includegraphics[width=\s,trim={80 0 80 30}, clip]{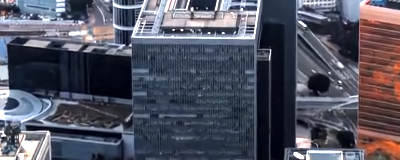} & \includegraphics[width=\s,trim={80 0 80 30}, clip]{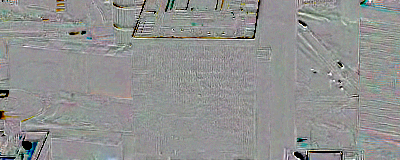} & \includegraphics[width=\s,trim={80 0 80 30}, clip]{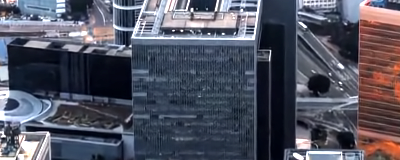}  
\end{tabular}
    \caption{Results of M2Mnet at a stack transition. We show the two last frames of one stack and the first frame of the next stack. Between them, we display the warping error.The contrast has been enhanced for better visualizatoin.}
    \label{fig:results_at_stack_transition}
\end{figure}

\myparagraph{Qualitative evaluation.}  Figure~\ref{fig:results_at_stack_transition} compares visual results of the baseline MIMO M2Mnet, the recurrent MIMO and our proposed recurrent OSO. 
We display the frames at time $t-2, t-1$ and $t$, a transition occurs from $t-1$ to $t$. We also show the difference between adjacent frames (aligned with~\cite{Perez2013}). This difference is similar for all the networks before the stack transition, and they are similar to the difference in the ground-truth. 
But at the stack transition, large inconsistencies can be observed for baseline and RAS, only the proposed ROSO is consistent as the ground-truth. In the supplementary material, we provide videos of these results in which the temporal inconsistency translates into low-frequency flickering. 
\color{black}

\section{Conclusion}


In this paper we discuss the MIMO approach, a recent trend in video restoration network architectures, focusing on the low-latency setting. We point out two problems in this setting: performance degradation within stacks and temporal discontinuity at stack transitions. We propose two solutions to address these problems: recurrence across stacks (RAS) and output stack overlapping (OSO). These two combined solutions can be applied to any MIMO network, and their effectiveness is demonstrated in this work for three networks from the state of the art: M2Mnet~\cite{chen2021multiframe}, BasicVSR++~\cite{chan2022generalization}, and ReMoNet~\cite{xiang2022remonet}. 
The proposed modifications thus define a new state of the art for low-latency applications.
Furthermore, the temporal inconsistency problem may go unnoticed on currently existing video benchmarks which feature strong, often shaky, motions between consecutive frames. To highlight this problem, we propose a new drone validation set with smooth motion in each sequence. We hope that this will encourage research towards better restoration on stabilized video.
Our evaluation of MIMO networks and proposed improvements is limited to denoising of additive white Gaussian noise. In principle, we expect our observations to carry over to other video restoration tasks (super-resolution, deblurring, deblocking) specially if the input is contaminated with temporally independent noise (such as Poisson noise, compressed noise), although this needs to be verified empirically.

\paragraph{Acknowledgments.} 
Work partially financed by DGA and FMJH PhD scholarships.
It was also performed using HPC resources from GENCI–IDRIS (grants 2023-AD011014015 and AD011011801R3) and from the “Mésocentre” computing center of CentraleSupélec and ENS Paris-Saclay supported by CNRS and Région Île-de-France.
Centre Borelli is also with Université Paris Cité, SSA and INSERM.

\bibliography{bibli}

\end{document}


\maketitle

\section{Videos for qualitative evaluation}

As part of the supplementary material, we provide four videos containing video denoising results obtained with M2Mnet, BasicVSR++ and ReMoNet, for AWGN with $\sigma=40$.
Each video shows in a split view the clean ground-truth, the noisy sequence, the results of the baseline networks and the results of ours baseline+ROSO (see the proposed method in Section 3 of the main paper). Some videos are played twice, the second time with enhanced contrast to highlight the differences between the methods.
All videos are encoded at 24 frames per second.


\begin{description}
\item[\texttt{drone-d160-40.mp4}] This is a crop from one of the videos of the proposed drone benchmark containing an urban scene.
The step-wise motion effect induced by the stack transitions can be noticed in low contrast textures (e.g. building facades), and also in the motion of the vehicles.
With the extensions proposed in the paper (+ROSO), the textures are much more stable and the cars on the road follow a more fluid motion. 
\item[\texttt{drone-R030-40.mp4}] The step-wise motion artifact is clearly visible on the texture of the threes in the foreground. On the other hand, they are stable in the reconstruction with our method. Note also the large change on the facades of houses and the cathedral.
\item[\texttt{REDS4-000-40.mp4}]
This is a sequence from the REDS dataset, taken with a hand-held GoPRO camera \cite{Nah_2019_CVPR_Workshops_REDS}. It features a strong camera shake.
The crop shows a region with a regular paved floor and grass. With the large amount of shake, the step-wise motion artifact is not noticeable.
\item[\texttt{REDS4\_stabilized-000-40.mp4}]
This time, we stabilized the sequence before denoising (using \cite{sanchez2017comparison,sanchez2018motion}). The step-wise motion artifact of the baseline low-latency MIMO networks becomes noticeable on this stabilized version of the sequence (step-wise fluctuations in the geometric texture of the paved floor, as well as in the grass). Again these changes are not visible with the proposed modifications.
This illustrates the fact that evaluating networks on videos with smooth motion is a crucial point in order to identify problems with temporal inconsistencies. 
\end{description}



\section{Networks and training details}\label{sec:networks}

We report results by applying the proposed modifications (recurrence across-stacks and output stack overlapping) to  three video denoising MIMO networks. These networks are all state of the art in the sense that they provide the best PSNR (and SSIM) among networks with a comparable computational cost. Thus, they stand at different ranges of the Pareto frontier depicted in Table 1 (left) of the main paper.
Below we provide a more detailed description of them.

\subsection{BasicVSR++~\cite{chan2022basicvsr++}}

This network was initially introduced for video super-resolution~\cite{chan2022basicvsr++} and later extended to video denoising and deblurring~\cite{chan2022generalization}. 
This MIMO network takes $N_i$ input frames and outputs $N_o=N_i$ frames. The size $N_i$ is determined by the GPU memory. The network first computes a feature representation for each input frame. This is done independently for each frame using residual convolutional blocks. 
It then processes these feature maps using a temporal fusion strategy which alternates between forward and backward propagation sweeps across the video segment. The propagation is  ``second-order'': for example, in the forward propagation sweep, when computing the output of layer $\ell$ at frame $t$, layer $\ell$ features from frames $t-2$ and $t-1$ are used. The propagation layers feature an alignment stage that registers the two previous feature maps for $t-2$ and $t-1$ to the one at frame $t$. This alignment is based on deformable convolutions~\cite{dai2017deformable}, guided by an optical flow~\cite{ranjan2017optical}.
The output frames are then decoded from the feature maps resulting from the propagation.

The original network uses this propagation inside a stack, but not across stacks. 
To produce a version of BasicVSR++ with  recurrence across stacks, we simply initialized the backward propagated features of the first two frames of the stack using the corresponding features of the last two frames of the preceeding stack. 
Note that the forward-propagated features of the last two frames of the stack are not available and thus are set to zero.

\paragraph{Training details.} We have not re-trained BasicVSR++, and used the weights provided by the authors.


\subsection{ReMoNet~\cite{xiang2022remonet}} 
To the best of our knowledge, ReMoNet~\cite{xiang2022remonet} is the only MIMO network in the video restoration literature with recurrence across stacks.
The code of ReMoNet is not available. We have implemented ReMoNet following the description provided in the paper.
ReMoNet~\cite{xiang2022remonet} follows a similar structure to FastDVDnet~\cite{tassano2020fastdvdnet}: two cascaded U-Nets considered as two denoising stages. The first U-Net is recurrent (called RTF in the paper). It takes as input three noisy frames plus the recurrent feature and outputs one frame and the updated recurrent feature. The second stage U-Net (called MOA) processes the $N_o$ intermediate frames, and outputs the final $N_o$ denoised frames. We use $N_o = 5$ output frames, which implies $N_i=7$ input frames. The recurrence of the first U-Net is not interrupted at stack transitions, thus this network is recurrent across stacks.

\paragraph{Training details.}
We trained several variants of the network for the ablation study. In all cases, we adapt the same training hyperparameters: we train 100 epochs with a batch size of 5 and 40,000 iterations per epoch. The initial learning rate is $10^{-4}$ and the learning rate is reduced by a factor of 10 at epoch 60 and 80. All models are trained for noise range 0-50.

\begin{figure}
    \centering
    \includegraphics[width=0.25\linewidth]{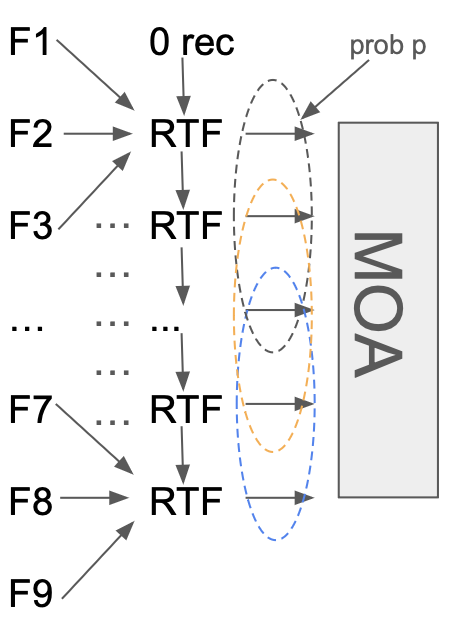}
    \caption{Training strategy used for ReMoNet. The RTF is recurrent. The MOA takes the outputs of 5 RTFs, which require a total of 7 frames to be produced. Instead, we load 9 frames and compute 7 RTF outputs. At random during training we select a number of warm-up RTFs (0, 1 or 2) whose outputs are discarded and use the following 5 RTF outputs. In this way we train the MOA on RTFs with different warm-ups. 
}
    \label{fig:train-stg-for-rec}
\end{figure}

During the training, 3D patches are cropped from training sequences. For non-recurrent versions, we only crop the exact number of frames that the network process in one evaluation, namely 7 frames. For the recurrent version we crop 3D patches with 9 frames.
The MOA takes 5 features produced by the 5 evaluations of the RTF. 
We apply 0, 1 or 2 ``warm-up'' RTFs to initialize the recurrence (the number of warm-up RTFs is chosen at random). The warm-up RTF features are discarded. Their purpose is to initialize the recurrence for the computation of the five RTF features for the 2nd stage MOA.
By doing this, we can avoid biaing the MOA to a specific RTF initialization, without increasing the training cost too much--we only run the MOA once. Figure~\ref{fig:train-stg-for-rec} shows our training strategy.
The second stage (MOA) must see a set of intermediate features where the first does not come from 0 recurrence feature map, otherwise it will perform very badly after a new stacks, due to the bias aforementioned. During the training, we crop a patch which consists of slightly more frames (9 or 10) than needed for one forward pass (7).

\subsection{M2Mnet~\cite{chen2021multiframe}} M2Mnet takes as input $N_i=7$ frames and outputs $N_o=7$ denoised frames. 
Throughout the entire network, the temporal dimension is preserved by encoding the video as a 4D tensor with spatial, temporal and channel dimensions. 
Two U-Nets process the video in parallel branches using 3D convolutions. One with only spatial 2D kernels (the \emph{intra-frame} network) and the other one with some 3D (separable) kernels (the \emph{inter-frame} network).
Their outputs are merged together to produce the final denoised frames. 
In M2Mnet, the authors use a loss term to enforce temporal consistency within the stack, which we did not find to be necessary: the network produces highly temporal consistent results \emph{inside} the stack without any temporal consistency loss. 

In the original M2Mnet is not recurrent. We implemented the network and a two recurrent variants, using the two options presented in Section 3.2 of the main paper: \emph{previous frame recurrence} or \emph{overlapped frames recurrence}. 
For the previous recurrence, we concatenate the previous denoised frame to the first input of the noisy stack. 
For the overlapped recurrence setting, we concatenate the $P$ output frames in the overlap region to the corresponding $P$ noisy input frames. 


\paragraph{Training details.}
For all the variants of M2Mnet, we train for 100 epochs with a batch size of 2 and initial learning rate of $4e-4$. We use the torch implementation of AdamW optimizer with the default parameters. During training, we use data augmentation made of transformations applied on the training batch (90 degrees rotations and reflections). 
The networks are trained for Gaussian noise with a standard deviation randomly sampled between $0$ and $50$.

To train the recurrent networks, we use batches made of clips with length $2 N_i$ (or alternatively $2 N_i - P$ for OSO). First of all, the network is applied on the first $N_i$ frames (first stack) and the recurrence feature-map is initialized with zeros. Then, the network is applied on the remaining frames (second stack) and the recurrence feature-map are taken from the denoised output of the first stack. We use a simple loss as the half sum of the two MSE losses computed respectively on the first and the second stack.

\section{Datasets}

In this section, we give more details about the training set we used.
As explained in the main paper (Section 4), we used two collections of datasets to train M2Mnet and ReMoNet. Firstly to produce the results on the drone benchmark (Table 1 of the main paper), we trained on a dataset of drone videos obtained from UAVs~\cite{li2016multi}, ERA~\cite{eradataset} and Visdrone~\cite{9573394}.\footnote{Note that the videos of the proposed drone benchmark were not obtained from any of these three drone datasets that we used for training.}
This large collection of training sets consists of 2963 sequences with a lot of variability.
From all sequences, we reserve 8 sequences for validation.
In the following, we briefly detailed the three datasets.

\noindent\textbf{UAVs}~\cite{li2016multi} is a dataset for target detection/tracking of \emph{Unmanned Aerial Vehicles}. Videos has been acquired with a GoPro3 HD camera mounted on a custom delta-wing airframe, either at resolution $1920 \times 1280$ or $1280 \times 960$. Most of them are around 50/60s at 30 fps

\noindent\textbf{ERA}~\cite{eradataset} is a benchmark for event recognition in aerial videos. This is a very large dataset made of 2864 aerial videos of roughly 5s at 24 fps. This dataset has a large diversity since it consists of 25 different classes. However, the sequences are low-resolution $640 \times 640$.

\noindent\textbf{Visdrone}~\cite{9573394} is a large-scale benchmark initially annotated for object recognition task, made of 63 videos at resolution $3840 \times 2160$, captured with various cameras and under various weather and lighting conditions.

\bigskip

For the results reported on standard benchmarks 
(Table 2 in the main paper), we retrained M2Mnet and ReMoNet (all variants) on a different training set, consisting in a collection of DeepVideoDeblurring~\cite{su17video} and REDS~\cite{Nah_2019_CVPR_Workshops_REDS}.
The DeepVideoDeblurring dataset consists in 71 videos of resolution $1280 \times 720$ or $1920 \times 1080$ taken with multiple cameras (iPhone 6s, GoPro Hero 4 Black, and Canon 7D).
For REDS, we follow~\cite{dewil2023video}, that is we temporally downsampled each sequence of the REDS-120 dataset~\cite{Nah_2019_CVPR_Workshops_REDS} by taking one frame over three. As the frame rate in REDS-120 is 120 fps, this leads to a dataset of videos with 40 fps. We keep the 240 training sequences of REDS, with 90 frames for each, at resolution $1280 \times 720$.
Since the videos in REDS-120 were acquired with a handheld camera, they contain large motion. As a data augmentation, we add the 240 REDS sequences after applying a video motion stabilization~\cite{sanchez2017comparison,sanchez2018motion} as in~\cite{dewil2023video}. Note that stabilizing the video creates black corners. In order to remove them, we follow~\cite{dewil2023video} and cropped the largest rectangle without black borders.

\section{Additional quantitative results}

\begin{table}
\centering
\scalebox{0.87}{
\begin{tabular}{ %
|c|c|c|c||c|c|c|c|c|}
    \hline
    \multirow{2}{3.5em}{Network} & \multirow{2}{.5em}{\rotatebox[origin=c]{90}{\footnotesize MIMO}} & \multirow{2}{.5em}{\rotatebox[origin=c]{90}{RAS}} & \multirow{2}{.5em}{\rotatebox[origin=c]{90}{OSO}} & \multicolumn{5}{c|}{PSNR/SSIM on drone} \\ 
    & & & & $\sigma=10$ &  $\sigma=20$ &  $\sigma=30$ &  $\sigma=40$ &  $\sigma=50$ \\ \hline
    \multirow{7}{3.5em}{M2Mnet} 
    &            &   &            &  40.59/.986 & 37.30/.972 & 35.27/.957 & 33.82/.942 & 32.70/.928 \\
    &            & P &            &  40.35/.986 & 37.09/.971 & 35.11/.956  & 33.69/.941  & 32.59/.927\\
    & \checkmark &   &            &  40,47/.986 & 37,09/.971 & 35,08/.956 & 33,65/.941 & 32,55/.927 \\
    & \checkmark & P &            &  40.22/.986 & 37.22/.973 & 35.38/.960 & 34.05/.947 & 33.00/.934 \\ \cdashline{2-9}
    & \checkmark &   & \checkmark &  40.81/\textbf{.987} &  37.44/.973 & 35.43/.959 & 34.00/.945 & 32.89/.931 \\
    & \checkmark & P & \checkmark &  \textbf{40.90/.987} & \textbf{37.74/.975} & 35.85/.963 & 34.50/.951 & 33.45./940 \\
    & \checkmark & O & \checkmark  &  40.75/\textbf{.987} &  37.71/\textbf{.975} & \textbf{35.86/.964} & \textbf{34.54/.952} & \textbf{33.50/.941} \\ \hline \hline                             
    \multirow{7}{4em}{\rotatebox[origin=c]{0}{ReMoNet}}
    &            &   &            &39.91/.985 &37.06/.972 &35.22/.958 &33.86/.944 &32.78/.930\\
    &            & P &            &40.02/\textbf{.986} &37.34/.974 &35.62/.962 &34.35/.951 &33.35/.939 \\
    & \checkmark &   &            &39.64/.984 & 36.67/.969 & 34.80/.954 & 33.45/.939 & 32.39/.924\\
    & \checkmark & P &            &40.20/\textbf{.986} & 37.35/.974 &35.58/.962 &34.29/.951 &33.28/.939\\ \cdashline{2-9}
    & \checkmark &   & \checkmark &39.82/.985  & 36.87/.970 & 35.02/.956 & 33.67/.941 & 32.62/.927 \\
    & \checkmark & P & \checkmark &\textbf{40.33}/\textbf{.986} &\textbf{37.48}/\textbf{.975} &\textbf{35.72}/\textbf{.963} &\textbf{34.44}/\textbf{.952} &\textbf{33.43}/\textbf{.941} \\ \hline \hline
    \multirow{3}{5em}{\rotatebox[origin=c]{0}{BasicVSR++}}  
    & \checkmark &   &            &  41.86/.989 & 38.62/.978 & 36.57/.967 & 35.09/.955 & 33.94/.943 \\
    & \checkmark & P &            & 42.13/\textbf{.990} & 39.09/.981 & 37.14/.971 & 35.71/.961 & 34.58/.951 \\\cdashline{2-9}
    & \checkmark & P & \checkmark & \textbf{42.41/.990} & \textbf{39.46/.982} & \textbf{37.53/.973} & \textbf{36.11/.963} & \textbf{34.98/.954} \\ 
    & \multicolumn{3}{c||}{full} & 43.06/.992 & 40.64/.986 & 39.01/.980 & 37.76/.974 & 36.73/.968 \\ \hline 
   Shift-net & \multicolumn{3}{c||}{low-latency} & 41.59/ .988 & 38.54/.978 & 36.57/.966 & 35.11/.954 & 33.99/.943 \\ 
   (plus) & \multicolumn{3}{c||}{full} & \textbf{42.35/.990} & \textbf{38.81/.983} & \textbf{38.09/.975} & \textbf{36.77/.967} & \textbf{35.71/.959} \\ \hline \hline
    \multirow{2}{2.5em}{RVRT} & \multicolumn{3}{c||}{low-latency} & 41.16/.988 & 38.24/.976 & 36.59/.964 & 35.51/.952 & 34.75/.940 \\ 
    & \multicolumn{3}{c||}{full} & 42.13/.990 & 39.75/.984 & 38.34/.977 & 37.34/.970 & 36.56/.964 \\ \hline \hline
    FastDVDnet & \multicolumn{3}{c||}{} & 39.23/.981 & 35.894/.962 & 33.93 /.944 & 32.55/.926 & 31.47/.909  \\ \hline
\end{tabular}}
\caption{Average PSNR/SSIM in our val-drone dataset, for Gaussian noise $\sigma=10,20,30,40$ and $50$. For each low-latency network, the values for the best variant are in bold.}
\label{tab:results_on_drone_dataset}
\end{table}

\begin{table}
\center
\scalebox{0.85}{
\begin{tabular}{|c|p{3.5cm}||p{1.5cm}|p{1.5cm}|p{1.5cm}|p{1.5cm} |p{1.5cm}|}
    \hline
    & \multirow{2}{4em}{Network}  & \multicolumn{5}{c|}{PSNR/SSIM on DAVIS-17} \\ 
    & & $\sigma=10$  & $\sigma=20$ & $\sigma=30$ & $\sigma=40$ & $\sigma=50$ \\  \hline
        \multirow{11}{1em}{{\rotatebox[origin=c]{90}{low latency}}}
    &M2Mnet (baseline)   &  39.34/.971 & 36.26/.946 & 34.45/.924  & 33.22/.903  & 32.25/.884  \\
    &M2Mnet (OSO prev rec)&  39.36/\textbf{.971} & 36.33/.947  & 34.55/.925  & 33.35/.906 & 32.40/.887 \\
    &M2Mnet (ours)        & \textbf{39.42/.971} & \textbf{36.42/.948} & \textbf{34.66/.927}  & \textbf{33.48/.908} & \textbf{32.53/.891}  \\ \cline{2-7}                           
    &ReMoNet (baseline)       &38.69/.967  &35.77/.941  &34.03/.917   &32.87/.897  &31.95/.878   \\
    &ReMoNet (ours)           &\textbf{38.75/.968} &\textbf{35.84/.942}  &\textbf{34.11/.918}  &\textbf{32.96/.898}   &\textbf{32.04/.879}    \\ \cline{2-7} 
    &BasicVSR++ (baseline)  & 40.33/.976 & 37.54/.959 & 35.81/.943  & 34.65/.928 & 33.68/.913  \\  
    &BasicVSR++ (ours)       & \textbf{40.64/.978} & \textbf{38.07/.964} & \textbf{36.45/.951}  & \textbf{35.37/.939} & \textbf{34.45/.926}  \\ \cline{2-7}   
    &Shift-net (plus)   & 40.29/.976 & 37.49/.959 & 35.84/.943 & 34.64/.927 & 33.70/.913 \\ \cline{2-7}  
    &RVRT   & 39.57/.972 & 37.31/.954 & 36.08/.937 & 35.32/.921 & 34.73/.905 \\   \cline{2-7}                   
    &FastDVDnet & 39.03/.968 & 35.94/.942 & 34.12/.918  & 32.91/.897 & 31.94/.877  \\
     \hline \hline
    \hspace{-.8em}\multirow{4}{1em}{{\rotatebox[origin=c]{90}{\shortstack{full\\ video}}}}

    &BasicVSR++\emph{/full}   & 40.80/.979 & 38.46/.967 & 37.00/.956  & 36.04/.947 & 35.18/937  \\ \cline{2-7} 
    &Shift-net\emph{/full} (plus) & 40.61/.977 & 38.07/.9638 & 36.59/.951  & 35.50/.940 & 34.62/.928 \\ \cline{2-7} 
    & RVRT\emph{/full} & 39.83/.974 & 37.86/.960 & 36.77/.947 & 36.1/.936 & 35.52/.924  \\ 
    \hline
\end{tabular}}
\caption{Average PSNR/SSIM/TC metrics for Gaussian noise ($\sigma=10,20,30,40,50$) on the DAVIS-17 dataset. For each low-latency network, the variant with the best performance is in bold. The networks at the top are low-latency, and their full versions are at the bottom where applicable.}
\label{tab:results_on_DAVIS}
\end{table}

\begin{table}
\center
\scalebox{0.85}{
\begin{tabular}{|c|p{3.5cm}||p{1.5cm}|p{1.5cm}|p{1.5cm}|p{1.5cm} |p{1.5cm}|}
    \hline
    & \multirow{2}{4em}{Network}  & \multicolumn{5}{c|}{PSNR/SSIM on Set8} \\ 
    & & $\sigma=10$  & $\sigma=20$ & $\sigma=30$ & $\sigma=40$ & $\sigma=50$ \\  \hline
        \multirow{11}{1em}{{\rotatebox[origin=c]{90}{low latency}}}
    &M2Mnet (baseline)   & 36.38/.952 & 33.64/.920 & 32.05/.892  & 30.93/.867 & 30.06/.843  \\
    &M2Mnet (OSO prev rec)&36.42/.952  & 33.71/.921 & 32.15/.895  & 31.04/.871  & 30.19/.849 \\
    &M2Mnet (ours)        & \textbf{36.52/.953} & \textbf{33.85/.924} & \textbf{32.30/.899}  & \textbf{31.20/.876}  & \textbf{30.34/.854}  \\ \cline{2-7}                           
    &ReMoNet (baseline)       &35.53/.946 &33.23/.915 & 31.77/.888 & 30.71/.863 & 29.90/.840    \\
    &ReMoNet (ours)           &\textbf{35.58/.946}& \textbf{33.29/.916} &\textbf{31.84/.889 }& \textbf{30.79/.865} & \textbf{29.98/.843}    \\ \cline{2-7} 
    &BasicVSR++ (baseline)  &  37.30/.959 & 34.84/.935 & 33.31/.912  & 32.19/.891 & 31.31/.872  \\  
    &BasicVSR++ (ours)       & \textbf{37.58/.962} & \textbf{35.32/.942} & \textbf{33.90/.923}  & \textbf{32.85/.906} & \textbf{32.00/.890}  \\ \cline{2-7}   
    &Shift-net (plus)   & 37.24/.958 & 34.77/.934 & 33.25/.911 & 32.15/.890 & 31.29/.872 \\ \cline{2-7}  
    &RVRT   & 37.58/.959 & 35.59/.934 & 34.61/.911 & 33.96/.890  &  33.47/.871 \\   \cline{2-7}                   
    &FastDVDnet & 36.24/.950 & 33.52/.916 & 31.90/.887  & 30.76/.861  & 29.87/.837   \\
     \hline \hline
    \hspace{-.8em}\multirow{4}{1em}{{\rotatebox[origin=c]{90}{\shortstack{full\\ video}}}}

    &BasicVSR++\emph{/full} & 37.82/.964 & 35.76/.948 & 34.48/.933  & 33.52/.919  & 32.74/.906   \\ \cline{2-7} 
    &Shift-net\emph{/full} (plus) & 37.49/.961 & 35.26/.942 & 33.88/.924 & 32.87/.907 & 32.07/.891 \\ \cline{2-7} 
    &RVRT\emph{/full}  & 37.90/.963 & 36.17/.944 & 35.33/.927  & 34.75/.912 & 34.30/.897  \\ 
    \hline
\end{tabular}}
\caption{Average PSNR/SSIM/TC metrics for Gaussian noise ($\sigma=10,20,30,40,50$) on the Set8 dataset. For each low-latency network, the variant with the best performance is in bold. The networks at the top are low-latency, and their full versions are at the bottom where applicable.}
\label{tab:results_on_Set8}
\end{table}

\begin{table}
\center
\scalebox{0.85}{
\begin{tabular}{|c|p{3.5cm}||p{1.5cm}|p{1.5cm}|p{1.5cm}|p{1.5cm} |p{1.5cm}|}
    \hline
    & \multirow{2}{4em}{Network}  & \multicolumn{5}{c|}{PSNR/SSIM on REDS4} \\ 
    & & $\sigma=10$  & $\sigma=20$ & $\sigma=30$ & $\sigma=40$ & $\sigma=50$ \\  \hline
        \multirow{11}{1em}{{\rotatebox[origin=c]{90}{low latency}}}
    &M2Mnet (baseline)   & 38.19/\textbf{.963} & 34.94/.932 & 33.10/.905  & 31.81/.881 & 30.82/.858  \\
    &M2Mnet (OSO prev rec)&  38.19/\textbf{.963}& 34.94/.932 & 33.12/.905  & 31.84/.881  & 30.85/.859 \\
    &M2Mnet (ours)        & \textbf{38.22/.963} & \textbf{35.00/.933} & \textbf{33.19/.907}  & \textbf{31.92/.884} & \textbf{30.94/.862}  \\ \cline{2-7}                           
    &ReMoNet (baseline)       &37.78/.960 &34.56/.927& 32.74/.899& 31.48/.874& 30.51/.851   \\
    &ReMoNet (ours)           & 37.82/.960 &34.60/.928 &32.78/.900& 31.52/.875& 30.56/.852    \\ \cline{2-7} 
    &BasicVSR++ (baseline)  &  39.45/.972 & 36.82/.954 & 35.22/.938  & 34.04/.922  & 33.11/.907  \\  
    &BasicVSR++ (ours)       &  \textbf{39.74/.974} & \textbf{37.35/.960} & \textbf{35.89/.946}  & \textbf{34.79/.933} & \textbf{33.90/.920}   \\ \cline{2-7}   
    &Shift-net (plus) & 39.00/.969 & 36.26/.949 & 34.66/.930 & 33.51/.913 & 32.60/.898 \\ \cline{2-7}  
    &RVRT   & 39.05/.969 & 36.59/.949 & 35.38/.930 & 34.62/.913  &  34.07/.897 \\   \cline{2-7}                   
    &FastDVDnet & 37.83/.959 & 34.55/.926 & 32.71/.898 & 31.44/.873  & 30.45/.849   \\
     \hline \hline
    \hspace{-.8em}\multirow{4}{1em}{{\rotatebox[origin=c]{90}{\shortstack{full\\ video}}}}

    &BasicVSR++\emph{/full}   & 39.85/.975 & 37.64/.963 & 36.31/.952  & 35.30/.941  & 34.47/.930   \\ \cline{2-7} 
    &Shift-net\emph{/full} (plus) & 39.19/.970 & 36.63/.952 & 35.13/.937 & 34.04/.922 & 33.19/.909 \\ \cline{2-7} 
    &RVRT\emph{/full} & 39.34/.972 & 37.09/.955 & 35.97/.940  & 35.23/.927 & 34.68/.914  \\ 
    \hline
\end{tabular}}
\caption{Average PSNR/SSIM/TC metrics for Gaussian noise ($\sigma=10,20,30,40,50$) on the REDS4 dataset. For each low-latency network, the variant with the best performance is in bold. The networks at the top are low-latency, and their full versions are at the bottom where applicable.}
\label{tab:results_on_REDS4}
\end{table}

\begin{table}
\center
\scalebox{0.85}{
\begin{tabular}{|c|p{3.5cm}||p{1.5cm}|p{1.5cm}|p{1.5cm}|p{1.5cm} |p{1.5cm}|}
    \hline
    & \multirow{2}{4em}{Network}  & \multicolumn{5}{c|}{PSNR/SSIM on REDS4-stabilized} \\ 
    & & $\sigma=10$  & $\sigma=20$ & $\sigma=30$ & $\sigma=40$ & $\sigma=50$ \\  \hline
        \multirow{11}{1em}{{\rotatebox[origin=c]{90}{low latency}}}
    &M2Mnet (baseline)   & 39.05/.971 &  35.97/.949 & 34.18/.929  & 32.90/.909  & 31.89/.891  \\
    &M2Mnet (OSO prev rec)& 39.06/.971 & 36.02/.950 & 34.26/.930  & 33.00/.911  & 32.00/.894  \\
    &M2Mnet (ours)        & \textbf{39.11/.972} & \textbf{36.11/.951} & \textbf{34.38/.932}  & \textbf{33.14/.914} & \textbf{32.16/.898}  \\ \cline{2-7}                           
    &ReMoNet (baseline)       &38.50/.968  &35.373/.943 &33.61/.921 &32.37/.901& 31.41/.883   \\
    &ReMoNet (ours)           &38.55/.968 &35.45/.944 &33.70/.922 &32.46/.903& 31.51/.885    \\ \cline{2-7} 
    &BasicVSR++ (baseline)  & 39.87/.976 & 37.19/.961 & 35.59/.947  & 34.40/.933 & 33.45/.920  \\  
    &BasicVSR++ (ours)       & \textbf{40.14/.978} & \textbf{37.71/.966} & \textbf{36.25/.955}  & \textbf{35.16/.944} & \textbf{34.27/.933}  \\ \cline{2-7}   
    &Shift-net (plus) & 39.77/.976 & 37.12/.960 & 35.53/.946 & 34.36/.933 & 33.43/.920 \\ \cline{2-7}  
    &RVRT   & 39.63/.975 & 37.13/.959 & 35.84/.944 & 35.00/.930 & 34.38/.916  \\   \cline{2-7}                   
    &FastDVDnet & 38.55/.968 & 35.44/.943 & 33.66/.921  & 32.39/.900 & 31.40/.881  \\
     \hline \hline
    \hspace{-.8em}\multirow{4}{1em}{{\rotatebox[origin=c]{90}{\shortstack{full\\ video}}}}

    &BasicVSR++\emph{/full}   & 40.26/.979 & 38.01/.969 & 36.71/.960  & 35.72/.951 & 34.91/.942  \\ \cline{2-7} 
    &Shift-net\emph{/full} (plus) & 40.03/.977 & 37.65/.965 & 36.25/.954 & 35.20/.944 & 34.35/.934 \\ \cline{2-7} 
    & RVRT\emph{/full} & 39.94/.977 &  37.71/.965 & 36.56/.954 & 35.78/.944  & 35.19/.934   \\ 
    \hline
\end{tabular}}
\caption{Average PSNR/SSIM/TC metrics for Gaussian noise ($\sigma=10,20,30,40,50$) on the REDS4-stabilized dataset. For each low-latency network, the variant with the best performance is in bold. The networks at the top are low-latency, and their full versions are at the bottom where applicable.}
\label{tab:results_on_REDS4_stabilized}
\end{table}

\paragraph{PSNR/SSIM. }In Table 1 (right) of the main paper, we provide the PSNR and SSIM of the different proposed variants of M2Mnet, ReMonet and BasicVSR++, on average on the drone benchmark and on average for the five noise levels $\sigma=10,20,30,40$ and $50$. In Table~\ref{tab:results_on_drone_dataset} of this supplementary material, we detailed the average PSNR and SSIM on the whole dataset for each individual noise level. We also include in this Table~\ref{tab:results_on_drone_dataset} the quantitative results of other networks from the literature which were used in the main paper: Shift-net (low-latency and full version), RVRT (low-latency and full version) and FastDVDnet.

Similarly, the Tables~\ref{tab:results_on_DAVIS}, \ref{tab:results_on_Set8}, \ref{tab:results_on_REDS4}, \ref{tab:results_on_REDS4_stabilized} show the detailed average PSNR and SSIM for these five noise levels on respectively DAVIS-17, Set8, REDS4 and REDS4-stabilized for all the tested networks (as in Table 2 of the main paper, which shows the average across the five noise levels).

\paragraph{Overlapped \emph{vs} previous output RAS.} For each benchmark dataset, we add in the tables~\ref{tab:results_on_DAVIS}, \ref{tab:results_on_Set8}, \ref{tab:results_on_REDS4},  \ref{tab:results_on_REDS4_stabilized} the comparison for M2Mnet between the ROSO with previous frame recurrence (referred as ``prev rec'') and the overlapped recurrence (denoted ``ours''). As explained in the main paper, the two options lead to similar results in the drone benchmark, but the overlapped recurrence is slightly better on other benchmarks. This might be due to the fact that the recurrent outputs in the overlapped recurrence are aligned with output frames in the current stack.
It makes sense that this option leads to a better performance, especially for strong motion, specially as M2Mnet does not contain explicit motion compensation.

\begin{figure}
    \centering
    \begin{tabular}{cc}
        \includegraphics[width=0.49\textwidth]{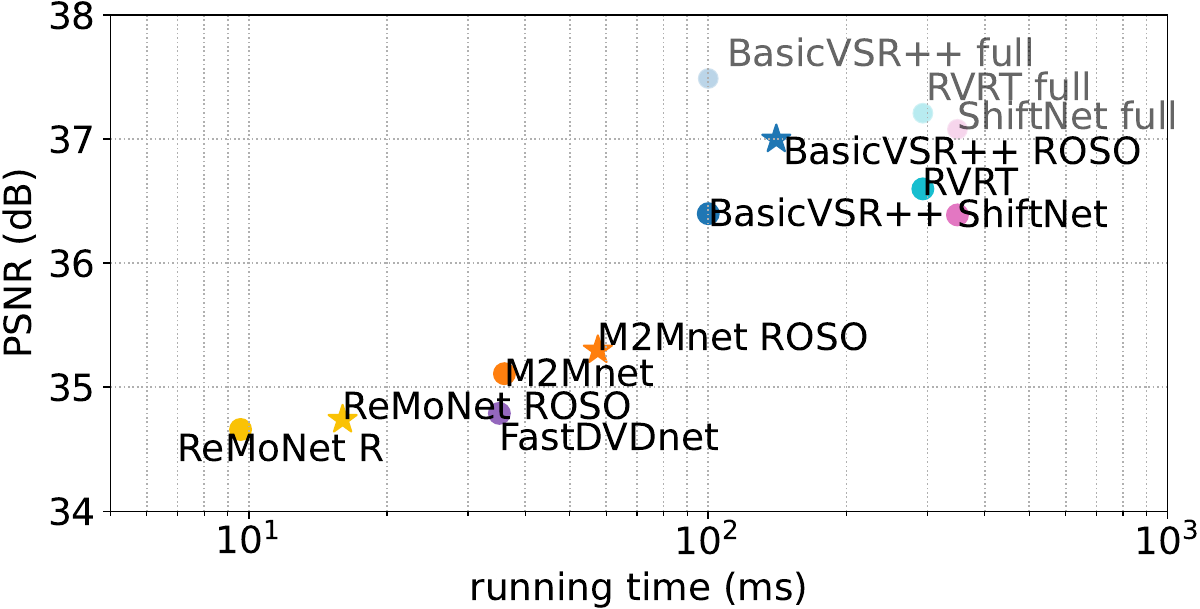} & \includegraphics[width=0.49\textwidth]{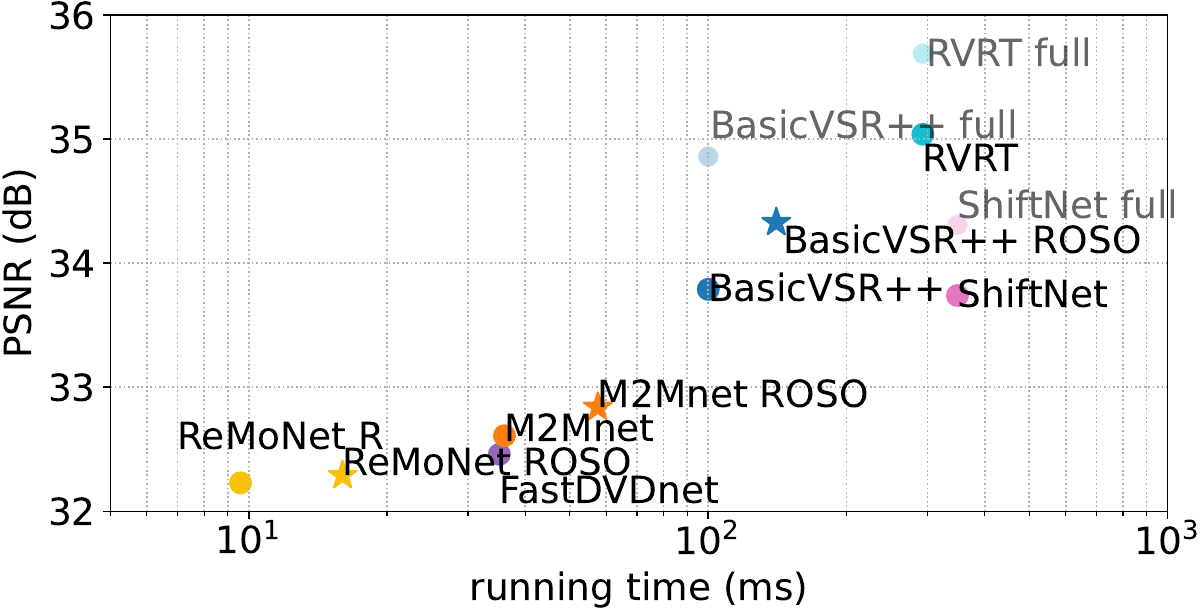}\\(a) DAVIS-17~\cite{Pont-Tuset_arXiv_2017} & (b) Set8~\cite{tassano2020fastdvdnet} \\ 
        \includegraphics[width=0.49\textwidth]{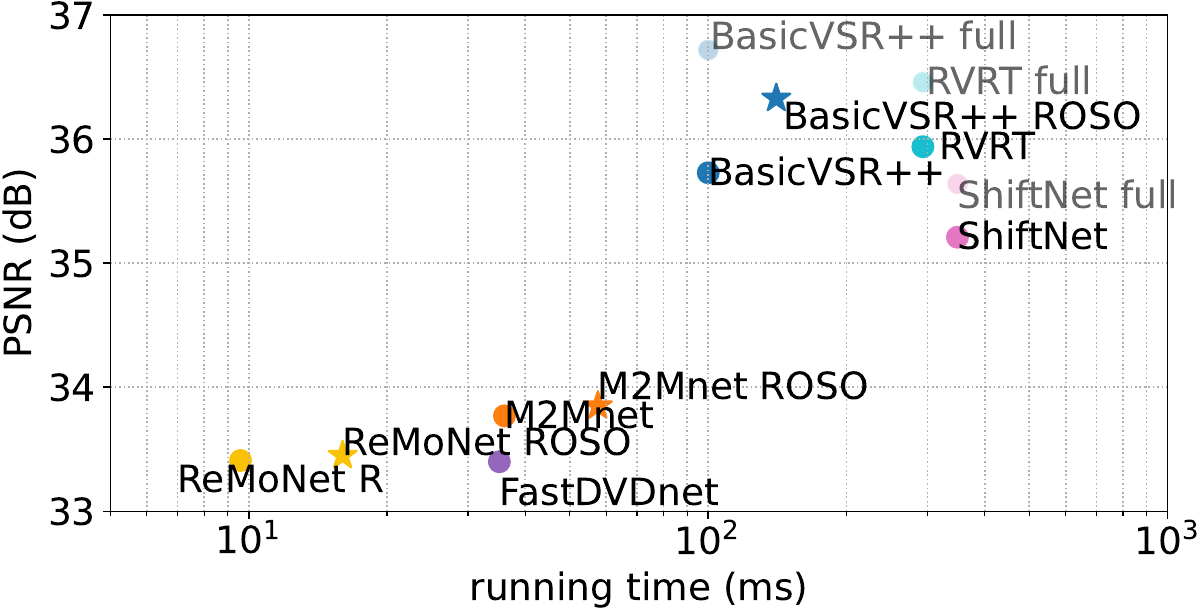} & 
    \includegraphics[width=0.49\textwidth]{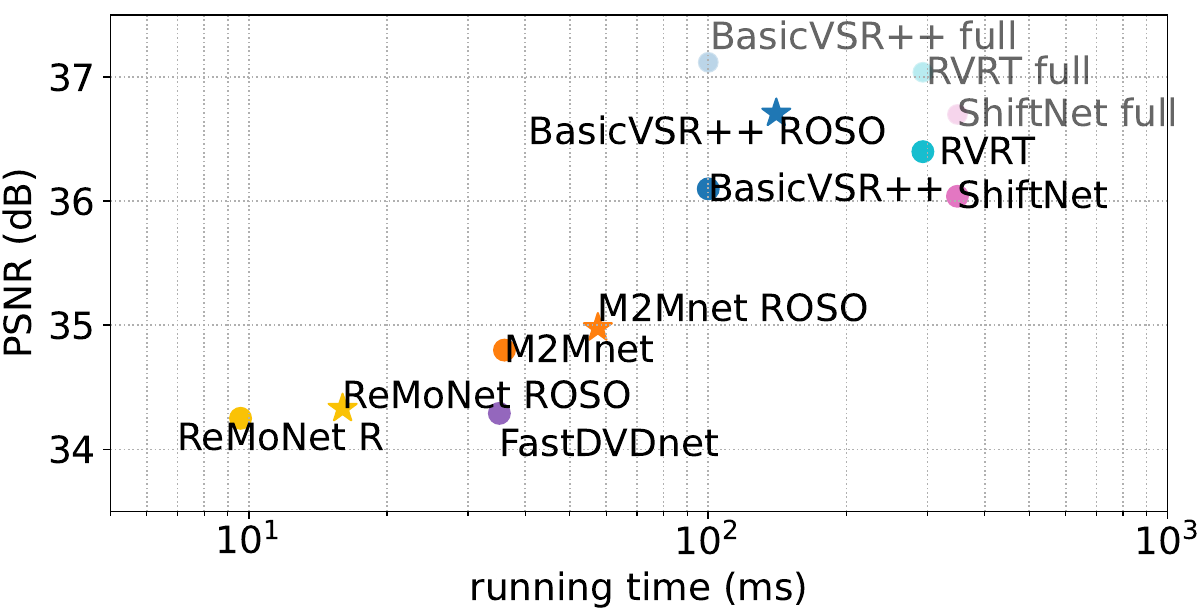} \\
     (c) REDS4~\cite{Nah_2019_CVPR_Workshops_REDS} & (d) REDS4-stabilized \\
     \includegraphics[width=0.49\textwidth]{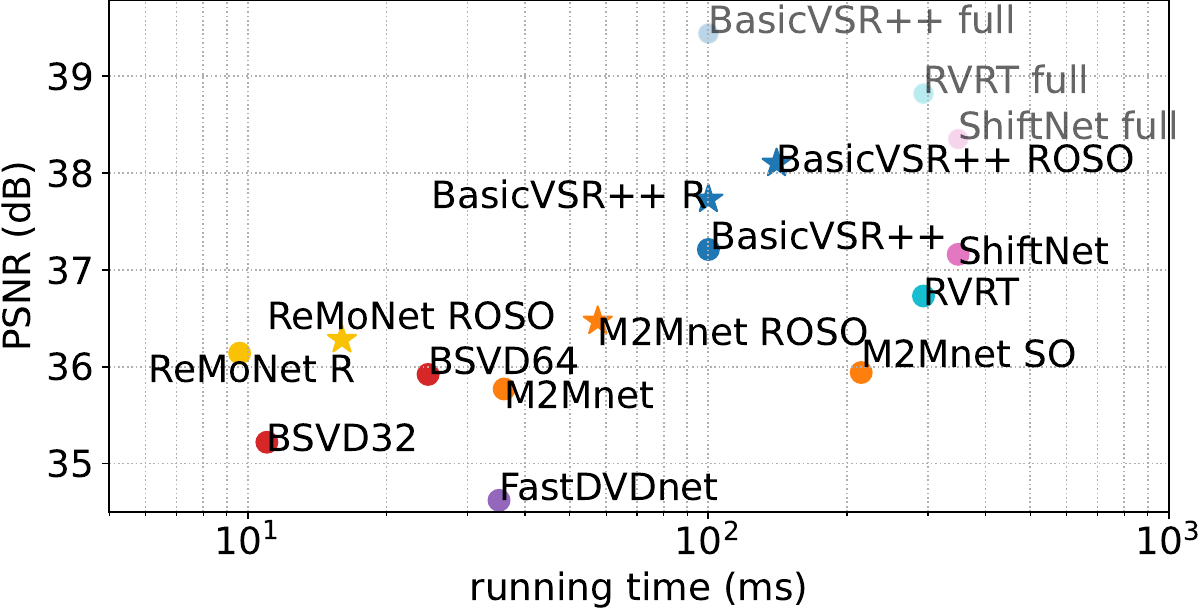} & \\
     (e) Drone benchmark &
    \end{tabular}
    \caption{Landscape of video denoising networks
derived from the results presented in the main papers. The vertical axis shows the averaged PSNR obtained over noise levels $\sigma=10,20,30,40,50$. The horizontal axis shows the running time per frame, measures on a Nvidia A100 GPU. The methods plotted with a star are variants proposed in this paper. As a reference, we added in lightgray the methods which takes as input the full video.}
    \label{fig:landscape}
\end{figure}

\paragraph{PSNR-runtime landscape.} In Figure~\ref{fig:landscape}, we show a landscape of PSNR vs running time of the methods tested in the paper, as in the Table 1 (left) of the main paper. The PSNR are measured in average over five noise levels $\sigma=10,20,30,40,50$ in the DAVIS-17 validation set (a), Set8 (b), REDS4 (c) REDS4-stabilized (d) and the drone benchmark (e).
For methods that can use a large MIMO stack, we use the low-latency version but also show as a reference the original version which takes as many frames as allowed by the GPU memory (denoted by ``full''). 
In (e), we display the same landscape as in the Table 1 (left) of the main paper, that is evaluated on the drone benchmark, but include for the comparison the performance of the full methods.

\begin{figure}[!ht]
    \centering
    \begin{tabular}{cc}
        \includegraphics[width=0.47\linewidth]{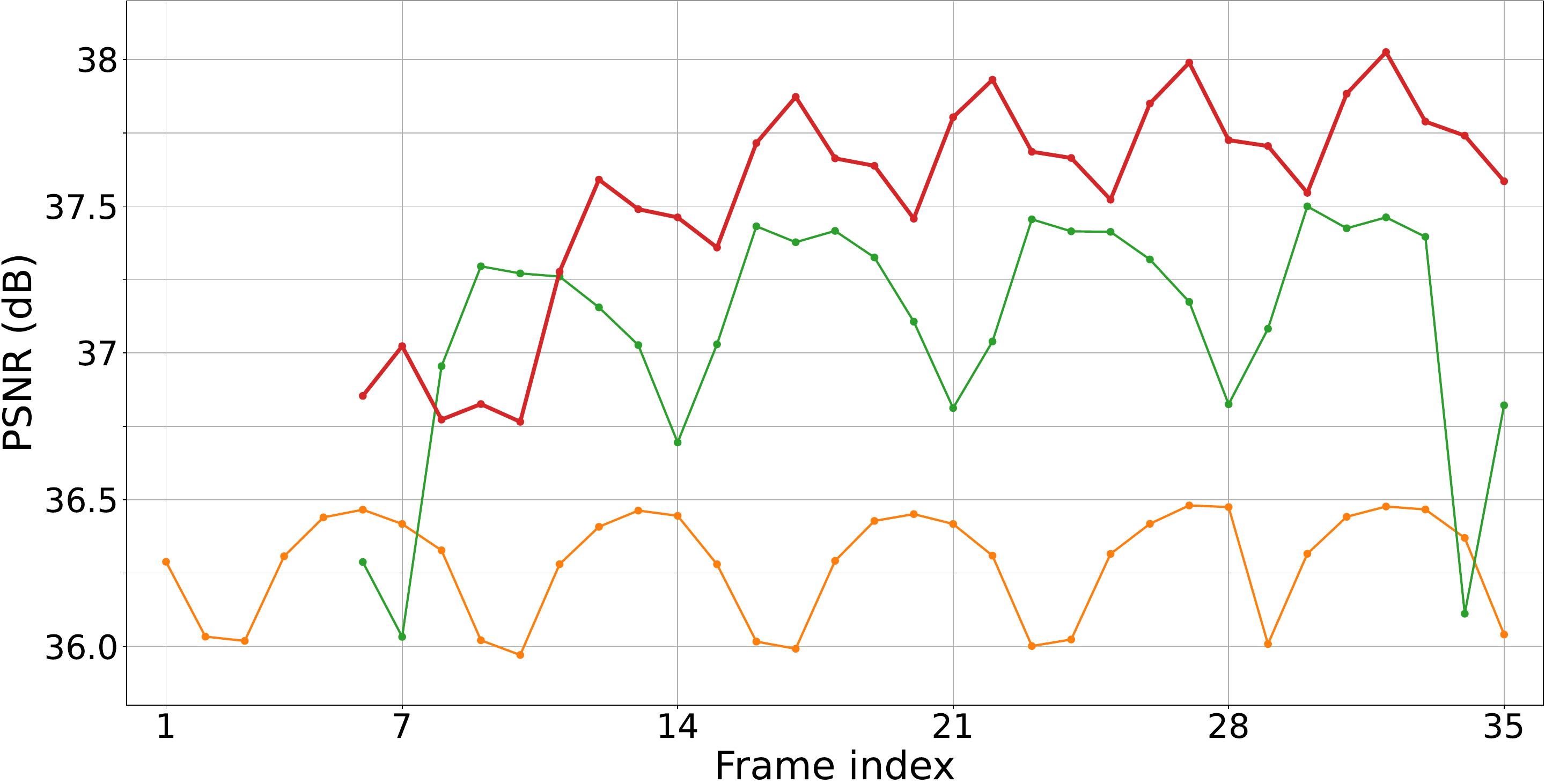} & \includegraphics[width=0.47\linewidth]{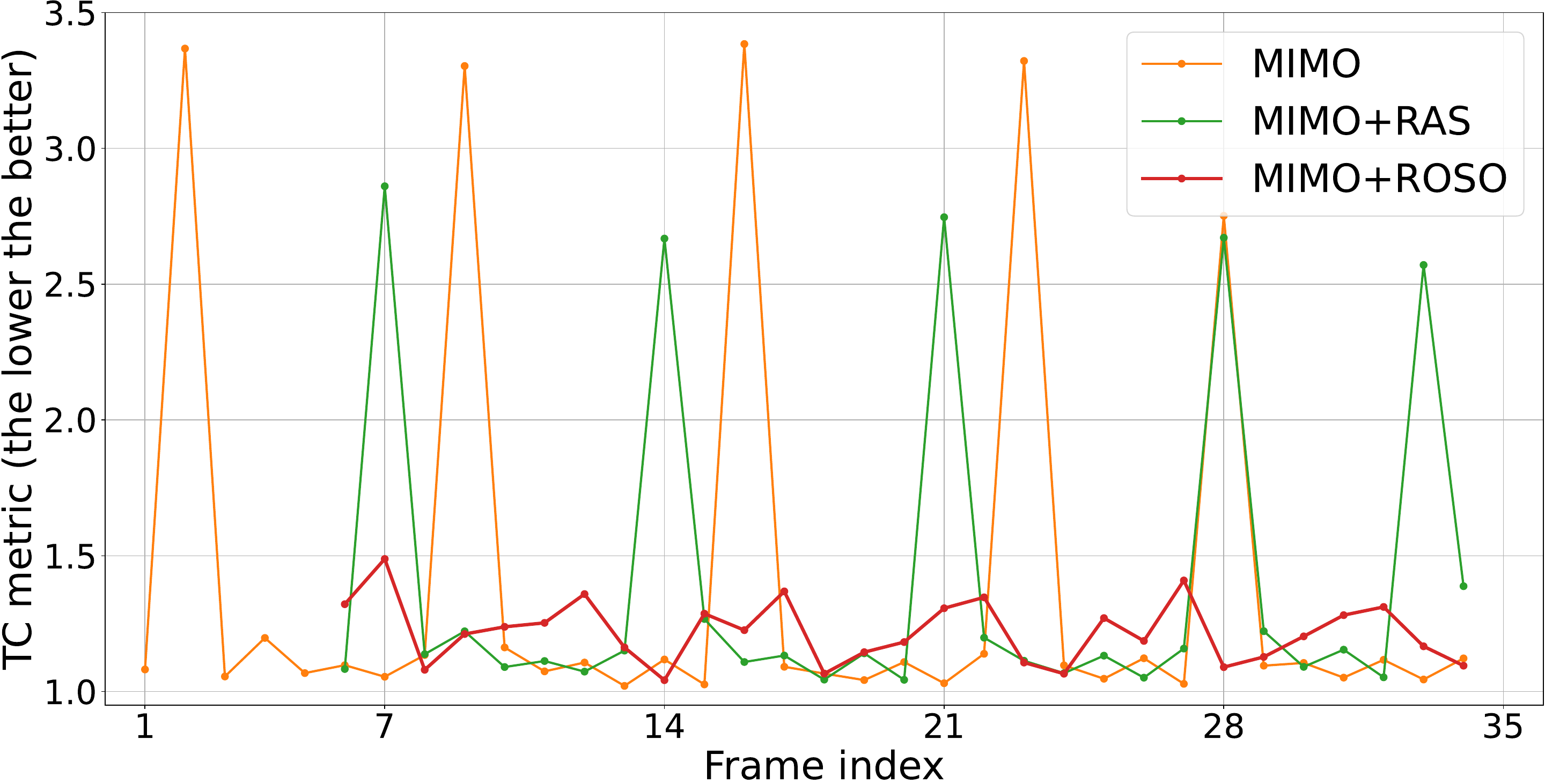}\\(a) PSNR per frame (BasicVSR++) & (b) TC per frame (BasicVSR++)
    \end{tabular}
    
    \vspace{0.5cm}
    
    \begin{tabular}{cc}
        \includegraphics[width=0.47\linewidth]{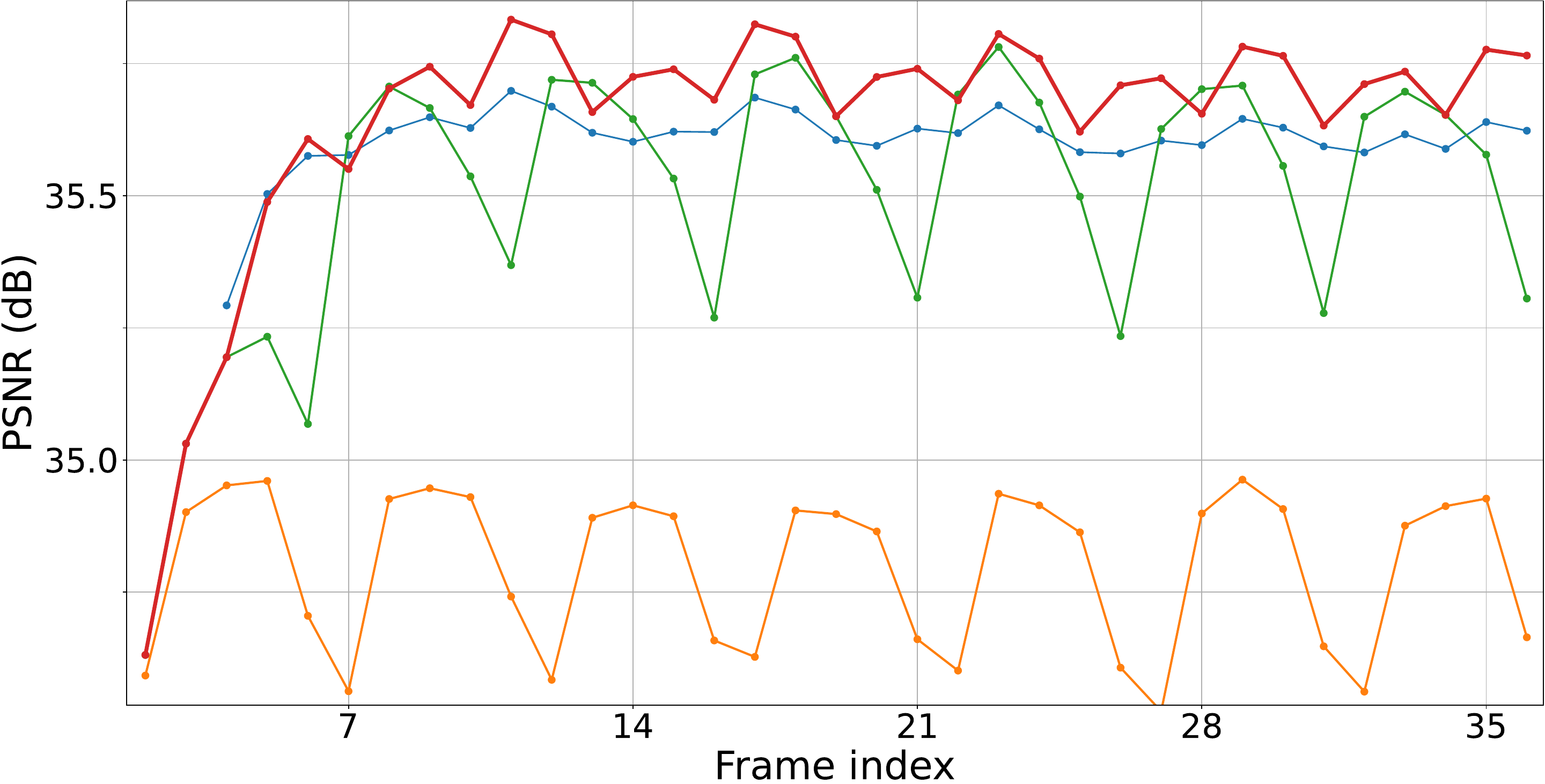} & \includegraphics[width=0.47\linewidth]{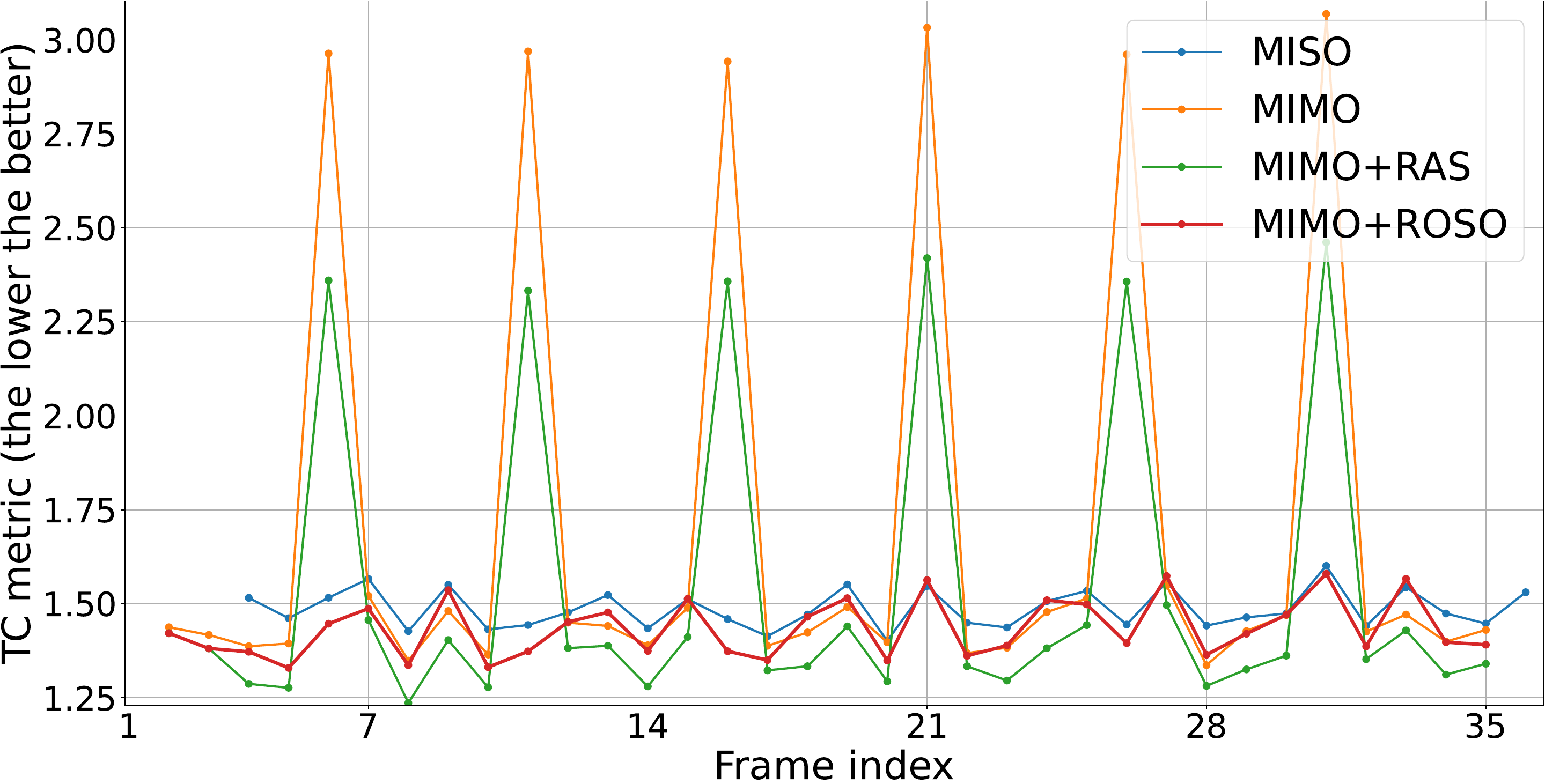}\\(c) PSNR per frame (ReMoNet) & (d) TC per frame (ReMoNet)
    \end{tabular}
    \vspace{-.5em}
    
  \caption{Per-frame PSNR and TC for BasicVSR++ and ReMoNet. For each frame we show the average PSNR  on the drone benchmark for AWGN denoising with $\sigma=30$.
  These curves show the same behaviour than the one of M2Mnet shown in Figure 1 of the main paper.}
    \label{fig:curves_PSNR_and_TC}
\end{figure}

\paragraph{PSNR and temporal consistency profile.} We complement Figure 1 in the main paper, by displaying the per-frame PSNR and TC profiles (Fig~\ref{fig:curves_PSNR_and_TC}) for respectively BasicVSR++ (top) and ReMoNet  (bottom) in average on the drone benchmark. 
The figures show that we obtain for BasicVSR++ and ReMoNet the same behaviour as for M2Mnet (Fig 1 in the main paper).

\section{Additional visual results}
\begin{figure}
\center
\setlength{\tabcolsep}{1pt} 
\def\s{0.19\textwidth}
\setlength{\tabcolsep}{1pt} 
\small
\def\seqzero{d160-026}
\def\seqone{d160-027}
\def\seqtwo{d160-028}
\def\seqthree{d160-029}
\begin{tabular}{ccccc}
     stack $N$; $t-2$ & & stack $N$; $t-1$ & $\leftarrow$\textbf{transition}$\rightarrow$  & stack $N+1$; $t$ \\
    \overimg[width=\s,trim={80 0 80 30}, clip]{figures/visual_results/gt-drone\seqone.png}{GT} & \includegraphics[width=\s,trim={80 0 80 30}, clip]{figures/visual_results/gt-drone\seqone-diff.png} &\includegraphics[width=\s,trim={80 0 80 30}, clip]{figures/visual_results/gt-drone\seqtwo.png} & \includegraphics[width=\s,trim={80 0 80 30}, clip]{figures/visual_results/gt-drone\seqtwo-diff.png} & \includegraphics[width=\s,trim={80 0 80 30}, clip]{figures/visual_results/gt-drone\seqthree.png} \\

    \vspace{0.25cm}
    
     \overimg[width=\s,trim={80 0 80 30}, clip]{figures/visual_results/noisy-drone\seqone.png}{noisy} & \includegraphics[width=\s,trim={80 0 80 30}, clip]{figures/visual_results/noisy-drone\seqone-diff.png} & \includegraphics[width=\s,trim={80 0 80 30}, clip]{figures/visual_results/noisy-drone\seqtwo.png} & \includegraphics[width=\s,trim={80 0 80 30}, clip]{figures/visual_results/noisy-drone\seqtwo-diff.png} & \includegraphics[width=\s,trim={80 0 80 30}, clip]{figures/visual_results/noisy-drone\seqthree.png}  \\
    
     \overimg[width=\s,trim={80 0 80 30}, clip]{figures/visual_results/M2Mnet_baseline-drone\seqone.png}{M2Mnet} & \includegraphics[width=\s,trim={80 0 80 30}, clip]{figures/visual_results/M2Mnet_baseline-drone\seqone-diff.png} & \includegraphics[width=\s,trim={80 0 80 30}, clip]{figures/visual_results/M2Mnet_baseline-drone\seqtwo.png} & \includegraphics[width=\s,trim={80 0 80 30}, clip]{figures/visual_results/M2Mnet_baseline-drone\seqtwo-diff.png} & \includegraphics[width=\s,trim={80 0 80 30}, clip]{figures/visual_results/M2Mnet_baseline-drone\seqthree.png}  \\

    \overimg[width=\s,trim={80 0 80 30}, clip]{figures/visual_results/M2Mnet_baselineRec-drone\seqone.png}{M2Mnet + RAS} & \includegraphics[width=\s,trim={80 0 80 30}, clip]{figures/visual_results/M2Mnet_baselineRec-drone\seqone-diff.png} & \includegraphics[width=\s,trim={80 0 80 30}, clip]{figures/visual_results/M2Mnet_baselineRec-drone\seqtwo.png} & \includegraphics[width=\s,trim={80 0 80 30}, clip]{figures/visual_results/M2Mnet_baselineRec-drone\seqtwo-diff.png} & \includegraphics[width=\s,trim={80 0 80 30}, clip]{figures/visual_results/M2Mnet_baselineRec-drone\seqthree.png} \\

    \vspace{0.25cm}
    
    \overimg[width=\s,trim={80 0 80 30}, clip]{figures/visual_results/M2Mnet_baselineRecOSO-drone\seqone.png}{M2Mnet + ROSO} & \includegraphics[width=\s,trim={80 0 80 30}, clip]{figures/visual_results/M2Mnet_baselineRecOSO-drone\seqone-diff.png} & \includegraphics[width=\s,trim={80 0 80 30}, clip]{figures/visual_results/M2Mnet_baselineRecOSO-drone\seqtwo.png} & \includegraphics[width=\s,trim={80 0 80 30}, clip]{figures/visual_results/M2Mnet_baselineRecOSO-drone\seqtwo-diff.png} & \includegraphics[width=\s,trim={80 0 80 30}, clip]{figures/visual_results/M2Mnet_baselineRecOSO-drone\seqthree.png}  \\

    \overimg[width=\s,trim={80 0 80 30}, clip]{figures/visual_results/BasicVSR_baseline-drone\seqtwo.png}{BasicVSR++} & \includegraphics[width=\s,trim={80 0 80 30}, clip]{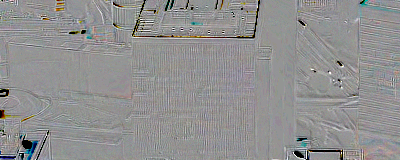} & \includegraphics[width=\s,trim={80 0 80 30}, clip]{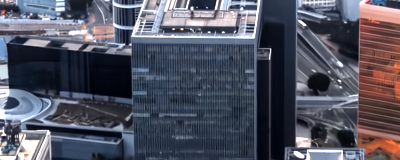} & \includegraphics[width=\s,trim={80 0 80 30}, clip]{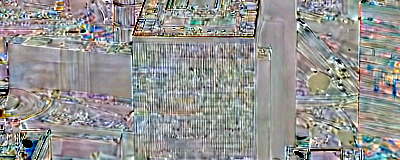} & \includegraphics[width=\s,trim={80 0 80 30}, clip]{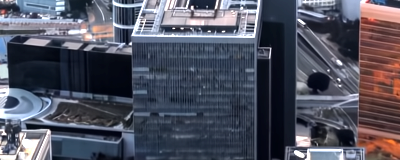}  \\

    \overimg[width=\s,trim={80 0 80 30}, clip]{figures/visual_results/BasicVSR_baselineRec-drone\seqtwo.png}{BasicVSR++ + RAS} & \includegraphics[width=\s,trim={80 0 80 30}, clip]{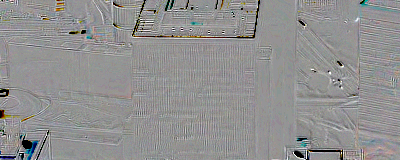} & \includegraphics[width=\s,trim={80 0 80 30}, clip]{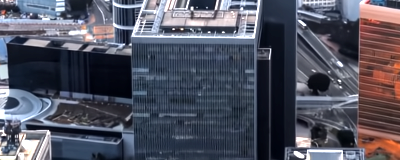} & \includegraphics[width=\s,trim={80 0 80 30}, clip]{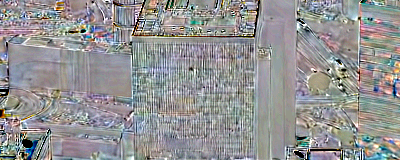} & \includegraphics[width=\s,trim={80 0 80 30}, clip]{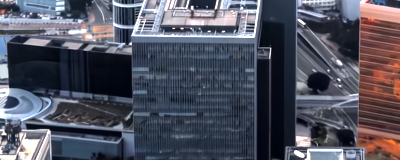} \\

    \vspace{0.25cm}
    
    \overimg[width=\s,trim={80 0 80 30}, clip]{figures/visual_results/BasicVSR_baselineRecOSO-drone\seqtwo.png}{BasicVSR++ + ROSO} & \includegraphics[width=\s,trim={80 0 80 30}, clip]{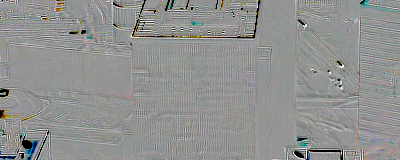} & \includegraphics[width=\s,trim={80 0 80 30}, clip]{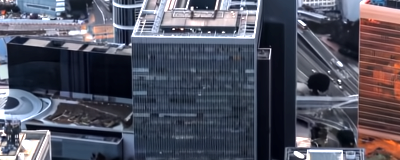} & \includegraphics[width=\s,trim={80 0 80 30}, clip]{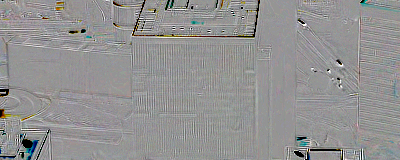} & \includegraphics[width=\s,trim={80 0 80 30}, clip]{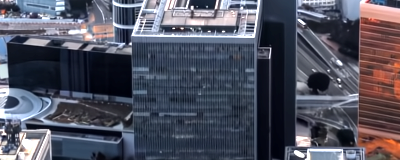}  \\

    \overimg[width=\s,trim={80 0 80 30}, clip]{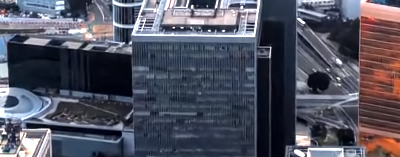}{ReMonet (==RAS)} & \includegraphics[width=\s,trim={80 0 80 30}, clip]{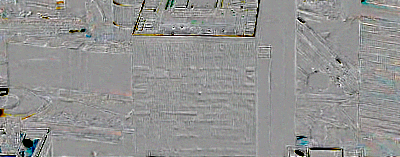} & \includegraphics[width=\s,trim={80 0 80 30}, clip]{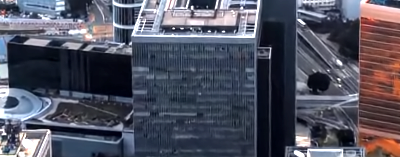} & \includegraphics[width=\s,trim={80 0 80 30}, clip]{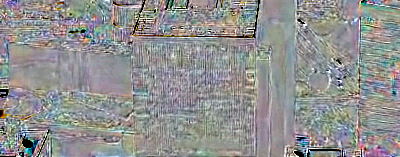} & \includegraphics[width=\s,trim={80 0 80 30}, clip]{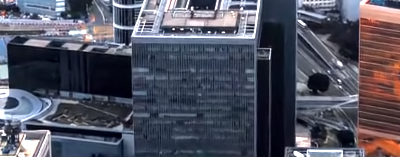}  \\

    \overimg[width=\s,trim={80 0 80 30}, clip]{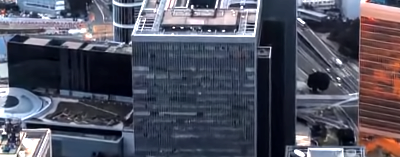}{ReMoNet + ROSO} & \includegraphics[width=\s,trim={80 0 80 30}, clip]{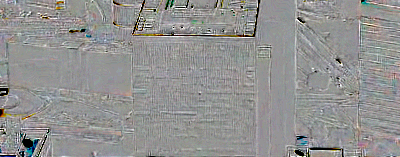} & \includegraphics[width=\s,trim={80 0 80 30}, clip]{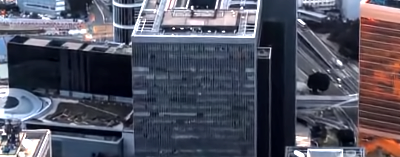} & \includegraphics[width=\s,trim={80 0 80 30}, clip]{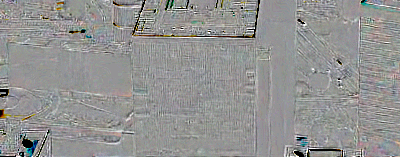} & \includegraphics[width=\s,trim={80 0 80 30}, clip]{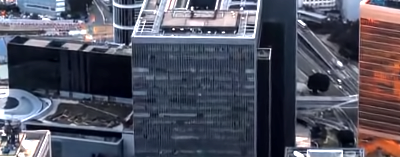}  
\end{tabular}

\def\seqone{R210-020}
\def\seqtwo{R210-021}
\def\seqthree{R210-022}

\def\seqone{C090-020}
\def\seqtwo{C090-021}
\def\seqthree{C090-022}
    \caption{Results of M2Mnet, BasicVSR++ and ReMoNet at a stack transition. For each, we compare the baseline network with the recurrent version (+ RAS) and the proposed method (+ ROSO). We show the two last frames of one stack and the first frame of the next stack. Between them, we display the difference (after aligning). For visualization purpose, the contrast has been enhanced. For the difference images, we map the range $[-27.7, 20]$ to [0,255]. }
    \label{fig:visual_results_d160_suppmat}
\end{figure}

\begin{figure}
\center
\setlength{\tabcolsep}{1pt} 
\def\s{0.19\textwidth}
\setlength{\tabcolsep}{1pt} 
\small
\def\seqzero{C090-026}
\def\seqone{C090-027}
\def\seqtwo{C090-028}
\def\seqthree{C090-029}
\begin{tabular}{ccccc}
     stack $N$; $t-2$ & & stack $N$; $t-1$ & $\leftarrow$\textbf{transition}$\rightarrow$  & stack $N+1$; $t$ \\
    \overimg[width=\s]{figures/visual_results/gt-drone\seqone.png}{GT} & \includegraphics[width=\s]{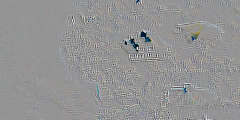} &\includegraphics[width=\s]{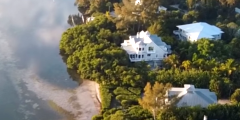} & \includegraphics[width=\s]{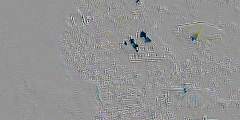} & \includegraphics[width=\s]{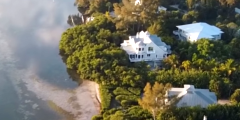} \\

    \vspace{0.25cm}
    
     \overimg[width=\s]{figures/visual_results/noisy-drone\seqone.png}{noisy} & \includegraphics[width=\s]{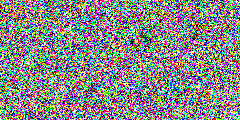} & \includegraphics[width=\s]{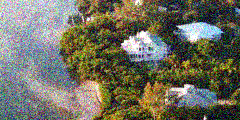} & \includegraphics[width=\s]{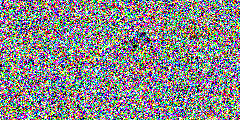} & \includegraphics[width=\s]{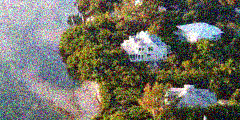}  \\
    
     \overimg[width=\s]{figures/visual_results/M2Mnet_baseline-drone\seqone.png}{M2Mnet} & \includegraphics[width=\s]{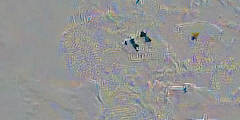} & \includegraphics[width=\s]{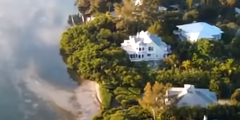} & \includegraphics[width=\s]{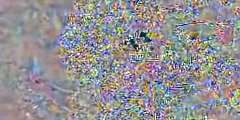} & \includegraphics[width=\s]{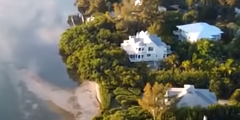}  \\

    \overimg[width=\s]{figures/visual_results/M2Mnet_baselineRec-drone\seqone.png}{M2Mnet + RAS} & \includegraphics[width=\s]{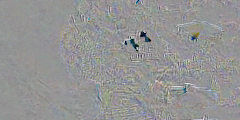} & \includegraphics[width=\s]{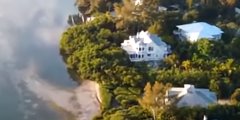} & \includegraphics[width=\s]{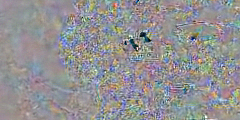} & \includegraphics[width=\s]{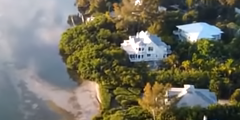} \\

    \vspace{0.25cm}
    
    \overimg[width=\s]{figures/visual_results/M2Mnet_baselineRecOSO-drone\seqone.png}{M2Mnet + ROSO} & \includegraphics[width=\s]{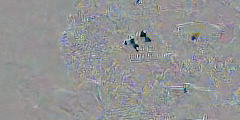} & \includegraphics[width=\s]{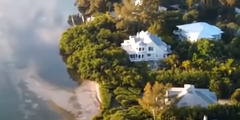} & \includegraphics[width=\s]{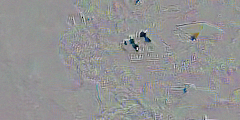} & \includegraphics[width=\s]{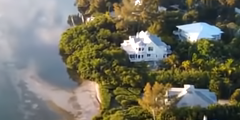}  \\

    \overimg[width=\s]{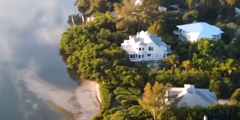}{BasicVSR++} & \includegraphics[width=\s]{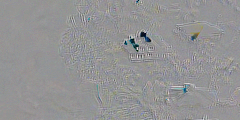} & \includegraphics[width=\s]{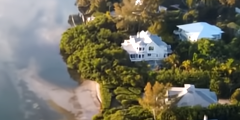} & \includegraphics[width=\s]{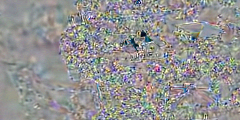} & \includegraphics[width=\s]{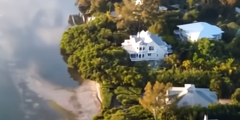}  \\

    \overimg[width=\s]{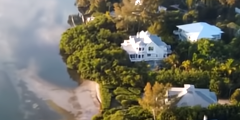}{BasicVSR++ + RAS} & \includegraphics[width=\s]{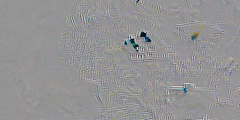} & \includegraphics[width=\s]{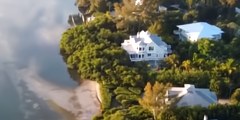} & \includegraphics[width=\s]{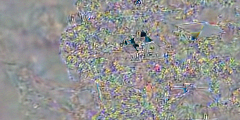} & \includegraphics[width=\s]{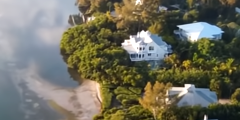} \\

    \vspace{0.25cm}
    
    \overimg[width=\s]{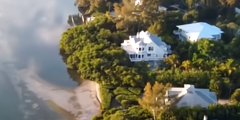}{BasicVSR++ + ROSO} & \includegraphics[width=\s,]{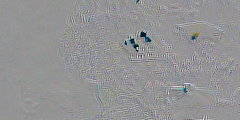} & \includegraphics[width=\s]{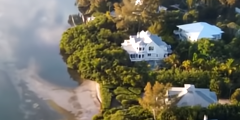} & \includegraphics[width=\s]{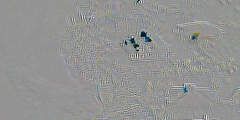} & \includegraphics[width=\s]{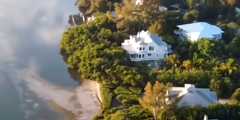}  \\

    \overimg[width=\s]{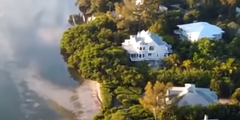}{ReMonet (==RAS)} & \includegraphics[width=\s]{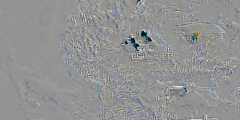} & \includegraphics[width=\s]{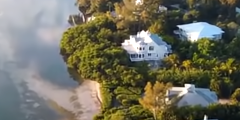} & \includegraphics[width=\s]{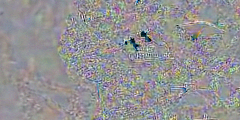} & \includegraphics[width=\s]{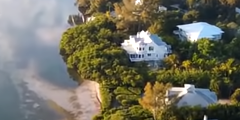}  \\

    \overimg[width=\s]{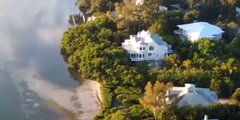}{ReMoNet + ROSO} & \includegraphics[width=\s]{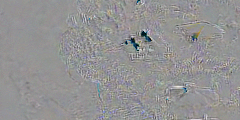} & \includegraphics[width=\s]{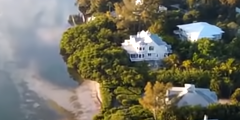} & \includegraphics[width=\s]{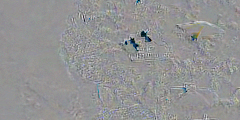} & \includegraphics[width=\s]{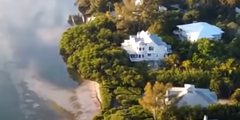}  
\end{tabular}
    \caption{Results of M2Mnet, BasicVSR++ and ReMoNet at a stack transition. For each, we compare the baseline network with the recurrent version (+ RAS) and the proposed method (+ ROSO). We show the two last frames of one stack and the first frame of the next stack. Between them, we display the difference (after aligning). For visualization purpose, the contrast has been enhanced. For the difference images, we map the range $[-27.7, 20]$ to [0,255].}
    \label{fig:visual_results_C090_suppmat}
\end{figure}

In Figures~\ref{fig:visual_results_d160_suppmat} and \ref{fig:visual_results_C090_suppmat}, we  provide more visual results of stack transition. Similarly to the Figure 3 in the main paper, we show the two last frames of one stack and the first frame of the next stack after transition, as well as the difference between two adjacent frames. For displaying the differences (computed between two images in the $[0,255]$ range), a contrast stretching has been applied. In these figures, we compare the baseline networks (non-recurrent, except for the ReMoNet) with the networks using recurrence across stacks, with and without output stack overlapping. We display the results for M2Mnet, but also BasicVSR++ and ReMonet.
It can be seen that all these three networks are very temporally consistent inside a stack. Indeed, the difference between frames within a stack is small and similar to the difference in the clean ground-truth. However, a large change occurs at stack transition for all of theses baseline networks. 
It is worth noting that using recurrence across stack only slightly attenuates the large change at stack transition. On the contrary, our propose ROSO spread the difference between multiple frames and hence the difference between frames in contiguous stack is largely attenuated.

\clearpage

\bibliography{bibli}